\documentclass{article}

\usepackage[nonatbib,preprint]{neurips_data_2022}

\usepackage[utf8]{inputenc} 
\usepackage[T1]{fontenc}    
\usepackage[colorlinks,linkcolor=black,citecolor=black]{hyperref}       
\usepackage{url}            
\usepackage{booktabs}       
\usepackage{amsfonts}       
\usepackage{nicefrac}       
\usepackage{microtype}      
\usepackage{xcolor}         
\usepackage{amssymb}
\usepackage{import}
\usepackage{subfigure}
\usepackage{wrapfig}
\usepackage{cite}
\usepackage{siunitx}
\usepackage[acronym,nonumberlist]{glossaries}
\usepackage{glossaries-prefix}
\newacronym{AD}{AD}{autonomous driving}
\newacronym{ADAS}{ADAS}{Advanced driver-assistance system}
\newacronym{AV}{AV}{automated vehicle}
\newacronym{AI}{AI}{artificial intelligence}
\newacronym{CMDP}{CMDP}{constrained Markov decision process}
\newacronym[shortplural=CNNs]{CNN}{CNN}{convolutional neural network}
\newacronym[shortplural=DTs]{DT}{DT}{decision tree}
\newacronym[shortplural=DQNs]{DQN}{DQN}{deep Q-network}
\newacronym{HRL}{HRL}{hierachical reinforcement learning}
\newacronym{IG}{IG}{integrated gradients}
\newacronym{IL}{IL}{Imitation Learning}
\newacronym[
    prefixfirst={a\ },
    prefix={an\ }
]{MCTS}{MCTS}{Monte Carlo tree search}
\newacronym[prefixfirst={a\ },prefix={an\ }]{MDP}{MDP}{Markov decision process}
\newacronym{ML}{ML}{machine learning}
\newacronym{PPO}{PPO}{proximal policy optimization}
\newacronym{RL}{RL}{reinforcement learning}
\newacronym{SMT}{SMT}{satisfiability modulo theories}
\newacronym{SUMO}{SUMO}{Simulation of urban mobility}

\usepackage{graphicx}
\usepackage{textcomp}

\title{
Driver Dojo: A Benchmark for Generalizable Reinforcement Learning for Autonomous Driving
}

\author{Sebastian Rietsch, Shih-Yuan Huang, Georgios Kontes, Axel Plinge, Christopher Mutschler\vspace{2mm}\\ 
{Fraunhofer IIS, Fraunhofer Institute for Integrated Circuits IIS, Nuremberg, Germany}\\
\begin{math}\{
    \begin{array}{c}
       \small{\texttt{sebastian.rietsch | huangsh | georgios.kontes}} \\
       \small{\texttt{axel.plinge | christopher.mutschler}}\\
    \end{array}\} \small{\texttt{@iis.fraunhofer.de}}
\end{math}
}%

\begin{document}

\maketitle
        
\begin{abstract}
\Gls{RL} has shown to reach super human-level performance across a wide range of tasks. However, unlike supervised machine learning, learning strategies that generalize well to a wide range of situations remains one of the most challenging problems for real-world \gls{RL}. \Gls{AD} provides a multi-faceted experimental field, as it is necessary to learn the correct behavior over many variations of road layouts and large distributions of possible traffic situations, including individual driver personalities and hard-to-predict traffic events. In this paper we propose a challenging benchmark for generalizable \gls{RL} for \gls{AD} based on a configurable, flexible, and performant code base. Our benchmark uses a catalog of randomized scenario generators, including multiple mechanisms for road layout and traffic variations, different numerical and visual observation types, distinct action spaces, diverse vehicle models, and allows for use under static scenario definitions. In addition to purely algorithmic insights, our application-oriented benchmark also enables a better understanding of the impact of design decisions such as action and observation space on the generalizability of policies. Our benchmark aims to encourage researchers to propose solutions that are able to successfully generalize across scenarios, a task in which current RL methods fail. The code for the benchmark is available at \url{https://github.com/seawee1/driver-dojo}.
\end{abstract}
\glsresetall

\section{Introduction}
\label{section:introduction}

Recently, \gls{RL} has exhibited impressive results in sequential-decision making problems that cannot be modeled with static decision rules or solved with classical optimization techniques, such as robotic manipulation~\cite{andrychowicz2020learning}, game-playing at superhuman-level~\cite{schrittwieser2020mastering} or designing next-generation chips~\cite{mirhoseini2021graph}. One of the key factors for these success stories was the combination of the recent advances in deep learning~\cite{goodfellow2016deep} with the core idea of the \textit{reward hypothesis}, which states that intelligence and its associated abilities can be understood as the maximization of reward~\cite{sutton2018reinforcement}. Thus, in contrast to classical methods requiring a formalization of the task performance into a complicated mathematical objective function, \gls{RL} offers a new angle to such problems.

This flexibility in the problem formulation becomes especially appealing in applications like \gls{AD}, where manually specifying all aspects of desired driving behavior for every possible emergent traffic configuration is impossible. The \gls{RL} paradigm in contrast makes it straightforward to specify objectives like \textit{arrive at the destination} or \textit{do not crash} through penalizing or rewarding the consequence of actions conditioned on the current world state.

Unfortunately this also comes at a cost: unlike the general understanding of the causes and best practices to address the bias-variance trade-off in supervised ML~\cite{goodfellow2016deep}, the complex interplay between overfitting in the training environment and generalization to unseen environments in \gls{RL} is not yet fully understood. The problem is amplified by evolving algorithms commonly being evaluated on benchmarks such as Atari or the ALE~\cite{bellemare2013arcade}, where generalization is not required as adapting to a specific environment and task is actually desired to underline the significance of algorithmic contributions.

Acknowledging this shortcoming, several \gls{RL} algorithms tailored towards robustness and generalizability have emerged~\cite{jiang2021prioritized,raileanu2021automatic,parker2022evolving}. A strong prerequisite of them is the availability of a pool of ``similar'' (training and evaluation) environments with varying difficulty or, alternatively, the ability to automatically generate them during training. This hinders the direct application of these algorithms to \gls{AD} as usually only non-standardized environments tailored to specific driving scenarios~\cite{highway-env} are available.

A notable effort in this direction is the adoption of real-world traffic datasets~\cite{zhan2019interaction, krajewski2018highd}, but the main problem here is that once the vehicle controlled by an \gls{RL} agent performs independent actions, the other traffic vehicles cannot dynamically react to this, since they follow the pre-recorded trajectories. What is required instead is the development of truly flexible, dynamic and extendable \gls{AD} benchmark environments that aim for evaluating the robustness and generalizability of \gls{RL} agents.

In this work, we develop a sophisticated, application-focused generalization benchmark for \gls{AD} (called Driver Dojo) to develop and compare new algorithms. Our contribution is a driving environment with the \gls{SUMO} engine as its backbone \cite{SUMO2018}. Driver Dojo offers a large suite of features: i) fully randomized street networks for intersections, roundabouts and multiple highway driving tasks; ii) fine-grained control over traffic initialization and randomization, including a sampling-based method for physical and behavioral non-ego driver attributes on a per-vehicle basis; iii) a direct and semantic action space, five different vehicle dynamics models, and a catalogue of ready-to-use observations; iv) an underlying modular and performant code-base that allows for a fine-grained composition of the environment. Furthermore, Driver Dojo offers a simple workflow for the creation and deployment of pre-defined scenarios, which can additionally be combined with the above mentioned traffic randomization techniques, while supporting a clean seeding mechanism which allows for full reproducibility~\cite{henderson2018deep}.

The remainder of this paper is organized as follows. Section \ref{section:background} provides background on (generalizable) RL and driving environments. Section \ref{section:dojo} introduces our Driver Dojo Environment. Section \ref{section:experiments} shows experiments and discusses results. Section \ref{section:conclusion} concludes.

\section{Background}
\label{section:background}

\subsection{Reinforcement learning}

A Markov Decision Process (MDP) is a mathematical framework that formalizes sequential decision-making problems. 
It is represented by a tuple \begin{math}\bigl< S,A,P,R,\gamma \bigr> \end{math}, where \begin{math}S\end{math} defines a set of states, \begin{math}A\end{math} represents a set of actions, \begin{math}P(s_{t+1} | s_t, a_t)\end{math} describes the state transition probabilities, \begin{math}R(s_t, a_t, s_{t+1})\end{math} is the reward function, and \begin{math}\gamma \in [0,1] \end{math} is the discount factor~\cite{sutton2018reinforcement}.  
At each time step \begin{math}t\end{math}, the agent obtains a representation of the world encoded in state \begin{math}s_t\end{math}, performs an action \begin{math}a_t\end{math} that is communicated to the environment, the environment transitions to a new state \begin{math}s_{t+1}\end{math} according to the transition model and provides the agent with a numerical reward \begin{math}r_t\end{math}. 
The goal of the agent it to learn a (deterministic or stochastic) optimal policy \begin{math}\pi^*(a_t | s_t)\end{math} that maps states \begin{math}s_t\end{math} to actions \begin{math}a_t\end{math} in a way that maximizes the cumulative future expected reward.
In cases where the agent cannot observe the complete world state $s_t$, we are confronted with a Partially Observable MDP (POMPD), and the agent is said to perceive an observation \begin{math}o_t \in \mathcal{O}\end{math} that represents an incomplete view of the world state.

There are several categories of Reinforcement Learning algorithms~\cite{arulkumaran2017deep} with distinct properties. We can differentiate between value-based and policy-based algorithms, with the former trying to estimate the cumulative future expected reward (also called value) of the current or of the optimal policy at any given state and the latter optimizing the policy directly. Actor-critic algorithms combine both worlds~\cite{szepesvari2010algorithms}. We can also distinguish between model-based algorithms~\cite{wang2019benchmarking}, which approximate the transition and the reward models from data~\cite{schrittwieser2020mastering} and utilize these approximators for faster policy learning or online planning, and model-free algorithms~\cite{duan2016benchmarking} that do not facilitate such intermediate steps. Finally, there exist \gls{RL} algorithms that require no interaction with the environment: Imitation Learning algorithms~\cite{osa2018algorithmic} utilize logged data from expert's demonstrations to mimic the (unknown) policy of the expert and Offline \gls{RL} algorithms~\cite{levine2020offline} that try to synthesize a policy that performs better compared to the policies used to collect the available data.

\subsection{Generalizable Reinforcement Learning}

In our benchmark, we mainly concern ourselves with i.i.d. generalization environments (as introduced by~\cite{DBLP:journals/corr/abs-2111-09794}), where train and evaluation scenarios are drawn from the same underlying distribution. However, as we show in our experiments, in-distribution generalization in an AD application already poses highly challenging problems.

Despite this plethora of \gls{RL} algorithmic advancements, it is widely understood that the training process is very sensitive not only to specific implementation choices but even to the random seeds of the (simulated) environments~\cite{henderson2018deep}. To add to this, a large body of work has recently shown that in several cases, the resulting trained policies can be brittle, as their performance degrades even with slight changes between the training and application environments~\cite{kirk2021survey}.
One of the simplest approaches to improve the robustness and generalization capabilities of trained policies is Domain Randomization (DR)~\cite{tobin2017domain}, but although it has been successfully applied to various difficult tasks, such as simulation-to-reality transfer~\cite{andrychowicz2020learning}, it can also fail for several cases~\cite{dennis2020emergent}.

(Automated) curriculum learning is an approach that can help agents to gain a foothold in complex environments and has also shown to have a positive impact on generalization~\cite{DBLP:journals/corr/abs-2011-08463}. We can also put a newly emerging category of algorithms named unsupervised environment design~\cite{dennis2020emergent,raileanu2021automatic,parker2022evolving} into this category. The lighter version, which solely arranges randomly encountered levels in a proper curriculum is PLR~\cite{jiang2021prioritized}.


\subsection{Reinforcement Learning for Autonomous Driving}


Recently, several algorithmic flavors of RL have been utilized to address various sub-problems in the autonomous driving domain~\cite{kiran2021deep}. In this direction, there have been notable successful real-world applications (e.g.~\cite{bojarski2016end, bansal2018chauffeurnet}) that leverage Imitation Learning algorithms~\cite{osa2018algorithmic} training on available data from vehicles, usually combined with some form of data augmentation to address the train/test data (and road environment) distribution shift. Other, smaller-scale studies utilize an available high-fidelity simulator and simulation-to-reality transfer approaches~\cite{osinski2020simulation} or even learn a driving policy online~\cite{kendall2019learning}.

A central concern in these approaches is to what extend we can \emph{guarantee} safety~\cite{shalev2017formal} for all possible situations that can occur in the real-world. To address different aspects of safety, complementary approaches have been adopted in the \gls{RL} community, stemming from designing more suitable observation and action spaces, by -- for example -- favoring feature learning over end-to-end control solutions~\cite{sauer2018conditional} and addressing high-level behavioral decisions (that carry semantic information) instead of controlling directly low-level actuators~\cite{shalev2016safe, mcallister2017concrete, rhinehart2021contingencies}, to incorporating safety and ``interpretability'' components directly on the training algorithm (e.g. see~\cite{kalweit2020deep, krasowski2020safe, schmidt2021trust, schmidt2022safe} and the references therein).

These stringent safety requirements, combined with the practically infinite real-world configurations that can emerge in every-day traffic, imply that the \gls{AD} use case can largely benefit from techniques that lead to robust and generalizable \gls{RL}-based driving agents.

\subsection{Driving Environments}

As there exists a large number of simulator and environments for autonomous driving we do not aim at providing a complete overview. Instead, we focus on the generalizability of benchmark suites and environments providing similar features as Driver Dojo. In particular, this includes MetaDrive~\cite{li2021metadrive} and ULTRA~\cite{Elsayed2020ULTRAAR}. For an exhaustive list of available environments and a comparison between them we refer the interested reader to the respective section in \cite{li2021metadrive}.

MetaDrive is one of the few frameworks that explicitly addresses generalizability and that focuses on composability of street network blocks, ranging from four-way and T-shaped intersections, to roundabouts and country roads, allowing to combine them into contiguous scenarios. In contrast to MetaDrive, we offer a smaller set of unique network constituents but instead introduce broader variations for these building blocks. Furthermore, we focus on realistic and diverse traffic modeling, whereas MetaDrive controls traffic with a single IDM traffic manager with fixed parameters.

Compared to ULTRA, which is based on the powerful SMARTS environment suite~\cite{zhou2020smarts}, Driver Dojo supports a much larger variety of intersection types, since ULTRA only specifically targets left-turn scenarios on T-shaped roads and cross-intersection. In addition, even though it introduces variations in the speed limit and traffic density, Driver Dojo has a more elaborate traffic modeling. ULTRA supports five basic vehicles including buses, trucks and trailers. While we did not yet implement them into Driver Dojo, these types are also available in SUMO and will be added in the near future.

\section{The Driver Dojo Environment}
\label{section:dojo}

\begin{figure}[t!]
\centering
\includegraphics[width=0.8\linewidth]{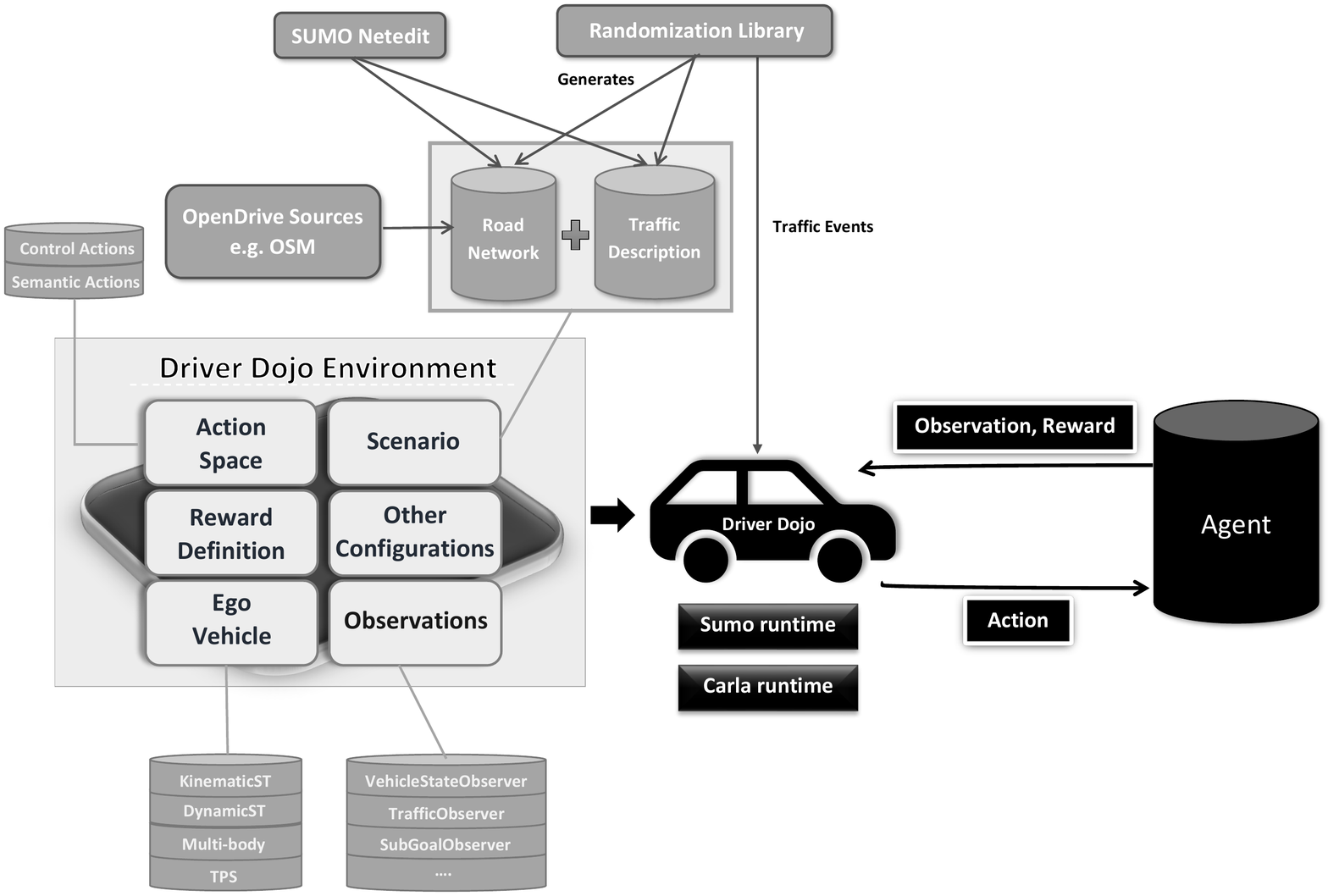}
\caption{Overview of Driver Dojo.}
\label{fig:greatness}
\end{figure}

Fig.~\ref{fig:greatness} gives an overview of the Driver Dojo environment and its interfaces. We will explain the open-source components that Driver Dojo is making use of (Section~\ref{subsection:dojo:1}), describe the action (Section~\ref{subsection:dojo:2}) and observation (Section~\ref{subsection:dojo:3}) spaces, the vehicle dynamics (Section~\ref{subsection:dojo:4}) and how we randomize the environments (Section~\ref{subsection:dojo:5}).

\subsection{Open-source Components}
\label{subsection:dojo:1}

\label{subsection:oss}
We employ SUMO as the core engine of Driver Dojo and use its traffic model to simulate traffic and its interactions with the ego vehicle.
It is a microscopic traffic simulator that is under active development for over 20 years and, thus, very mature and offers a pallet of features and a versatile ecosystem of support applications, making it attractive in many aspects for building a naturalistic driving benchmark for autonomous agents.
The driving simulator Carla~\cite{dosovitskiy2017carla} offers official SUMO co-simulation support in its recent versions.
Even though SUMO lacks a native 3D engine, the integration of Carla allows us to provide diverse sensor readings through an extensive library of implemented sensor models.
We take advantage of the vehicle dynamics library of CommonRoad~\cite{althoff2017commonroad} to simulate ego vehicle motion. It offers a range of dynamic models of varying complexity and physical parameter sets of three distinct real-world passenger cars.
Lastly, junction-art~\cite{junctionart:2023} and the scenariogeneration Python package are used to generate OpenDRIVE maps.

Developing and maintaining an AD environment is a lot of effort, as it is a challenging application that is technologically advancing and evolving rapidly.
Thus, we argue that exploiting available open-source solutions is key when developing an AD environment.
In our case, improvements made in SUMO will allow for more realistic traffic conditions, and progression in the development of, for example, the \textit{scenariogeneration} package will allow for realizing even broader distributions of street networks and driving scenarios.
This way, our Driver Dojo benchmark will hardly become stale over the course of the upcoming years.
To ensure comparability with results achieved on past versions, we will further employ proper versioning of our benchmark.


\subsection{Driving Actions}
\label{subsection:dojo:2}

Our implementation allows granting the agent \textit{direct} access to the car's throttle, brake pedal, and steering wheel or commanding it through high-level \textit{semantic} actions. 
Whereas control actions have an immediate and direct influence on the vehicle, semantic actions affect the vehicle many steps into the future, abstracting the underlying control complexity and, in other words, decoupling the behavioral execution problem from the planning problem, allowing to solve them independently. 
In its current form, the semantic action space is realized through a low-level Stanley control that tracks the path along a selected lane under a specified velocity, both of which the agent can modify through its actions. 
More formally, our semantic actions are a mapping $a_t, l_t, v_t \mapsto v_{t+1}, l_{t+1}$ where $l_t$ is the lane assignment, $v_t$ the velocity, and $a_t$ the action at timestep $t$. Lane-related actions are switching between lanes on normal roads, and choosing the lane to follow in the case of an upcoming junction.
As future work, we may think about additional semantic action spaces.

To ensure maximum flexibility, action spaces are parameterized, allowing, for example, to discretize the continuous control action space or choosing between different discrete space make-ups. Additionally, we offer an interface for outside motion planners through passing a list of waypoints to the agent vehicle, enabling combined methods of classical motion planning and RL.

\subsection{Observing the Road Situation}
\label{subsection:dojo:3}

It is common practice to either provide raw perceptions in the form of sensor readings to the agent or let him sense the environment in the form of driving affordances \cite{chenDeepDrivingLearningAffordance2015b,agarwalAffordancebasedReinforcementLearning2021}, which is a more condensed representation of the road situation inside a meaningful and compact feature space.
Whereas the latter eases the learning task, reward-based feature learning on raw perceptions could be an important angle for solving the generalization problem, as convolutional neural networks have proven immensely powerful in learning expressive feature representations in other problem domains.
Driver Dojo enables both paradigms by implementing a heap of ready-to-use ground-truth observations extracted from the world state inside SUMO and exploiting Carla to generate raw perception sensor signals. An overview of the available observations is given in Table~\ref{table:observers}.

\begin{table}
\caption{Observation types offered by Driver Dojo that can be freely combined with each other. The variables $n, m, h$ and $w$ depend on configuration. For the \texttt{RoadShape} observer, $n$ is the number of rays and $m$ is the number of intersections per ray. The \texttt{Carla} observation size highly depends on the sensor model configuration.}
\label{table:observers}
\centering
\vspace{4pt}
\begin{tabular}{lll}
\toprule
Observer     & Description     & Size \\
\midrule
\texttt{EgoState} & State of the ego vehicle & $6$ \\
\texttt{TrafficState} & Non-ego traffic state inside a fixed radius & $6n$\\
\texttt{RoadShape} & Ray-based road exterior sampling & $2nm$\\
\texttt{Navigation} & Waypoint and (sub-)goal information & $4n$ \\
\texttt{TrafficLight} & Upcoming traffic light state (one-hot) & $3$ \\
\texttt{RoadOptions} & Allowed semantic actions (one-hot) & $5$ \\
\texttt{BirdsEye} & Simplified, birds-eye scene rendering & $3 \times h \times w$ \\
\texttt{Carla} & Sensor-observations from Carla & variable\\
\midrule
\bottomrule
\end{tabular}
\end{table}

Observation query, assembly, and transformation logic are partitioned into separate observer classes, each dedicated to a particular environment aspect or context.
This design allows for a clean and fine-grained configuration of the observation space, which is especially helpful for building new research scenarios or optimizing the observation space as part of a hyperparameter search. 
At the same time, it minimizes overhead when implementing new observer classes and increases usability.
We mask the multiplicity of instantiated observers through a wrapper class that automatically collects and merges different observation fragments and supports pure vector or image-based observations and mixtures of both. 

\subsection{Vehicle Dynamics}
\label{subsection:dojo:4}

To decrease the computational complexity for experimental settings where accurate physical modeling is not the main concern, Driver Dojo offers the TargetPositionSpeed (TPS) vehicle model, in which we simply interpolating the ego position and orientation towards the next waypoint.

In all other cases, we adopt the kinematic single-track, single-track, single-track drift and multi-body dynamics models vehicle dynamics models offered by the CommonRoad~\cite{althoff2017commonroad} suite, along with the accompanying physical parameters for a Ford Escort, BMW320i, and a VW Vanagon.





\begin{figure}[t!]
\subfigure[Roundabout]{\hspace{.18\linewidth}\label{fig:my_roundabout}}\hfill
\subfigure[Intersection]{\hspace{.18\linewidth}\label{fig:my_intersection}}\hfill
\subfigure[Highway-Entry]{\hspace{.18\linewidth}\label{fig:my_highwayentry}}\hfill
\subfigure[Highway-Drive]{\hspace{.18\linewidth}\label{fig:my_highwaydrive}}\hfill
\subfigure[Highway-Exit]{\hspace{.18\linewidth}\label{fig:my_highwayexit}}

\subfigure{\includegraphics[width=.18\linewidth]{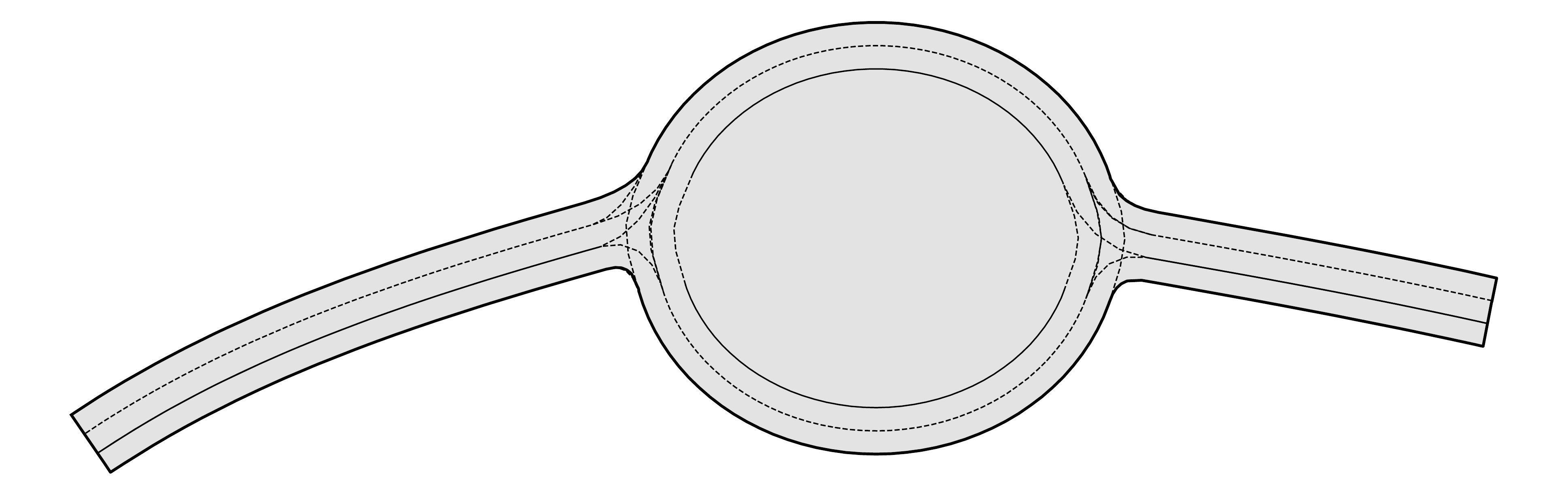}}\hfill
\subfigure{\includegraphics[width=.18\linewidth]{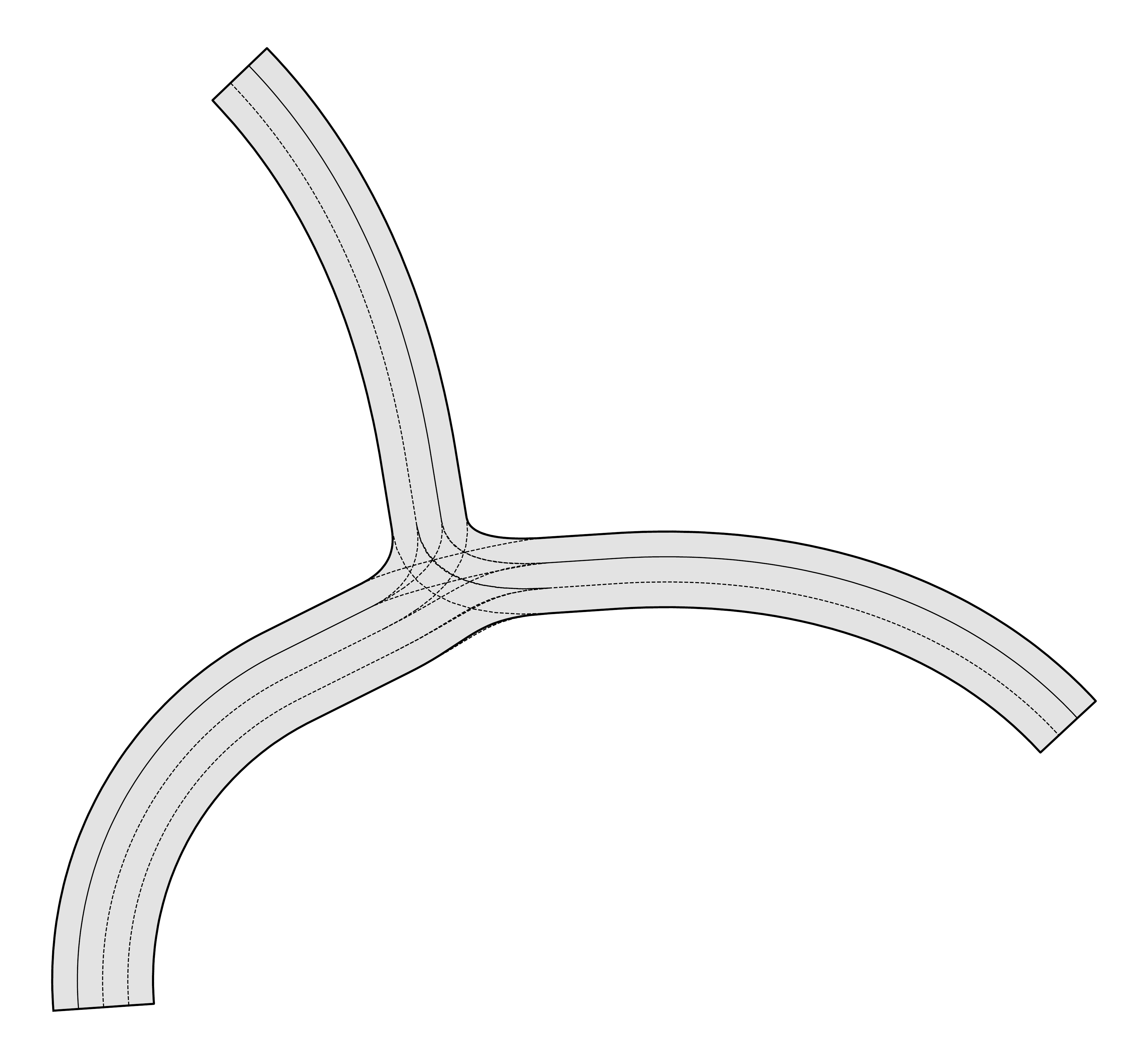}}\hfill
\subfigure{\includegraphics[width=.18\linewidth]{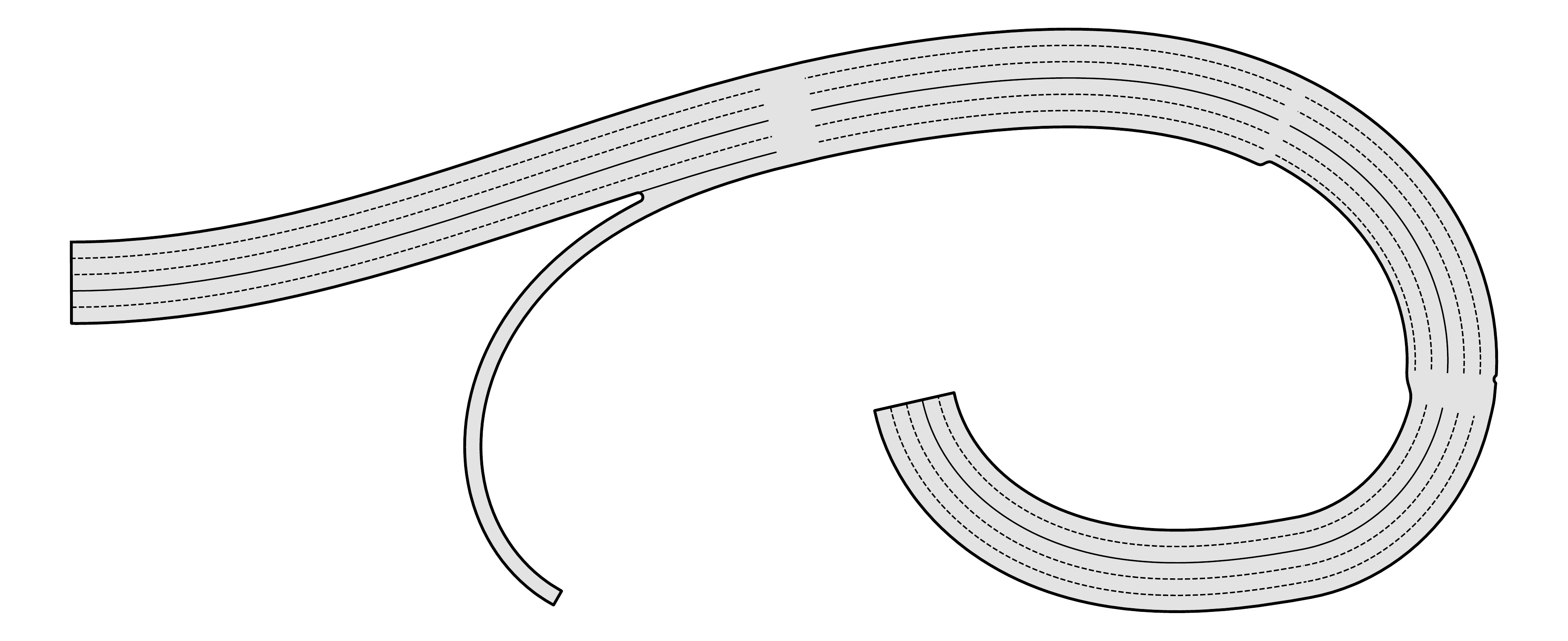}}\hfill
\subfigure{\includegraphics[angle=90,origin=c,width=.18\linewidth]{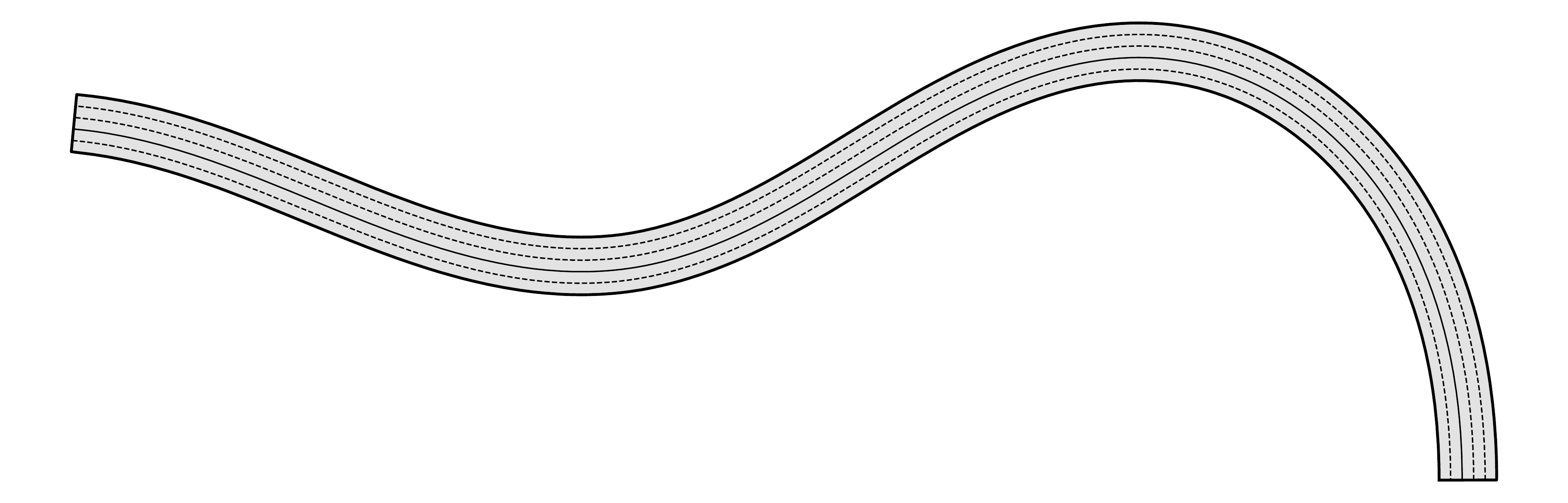}}\hfill
\subfigure{\includegraphics[width=.18\linewidth]{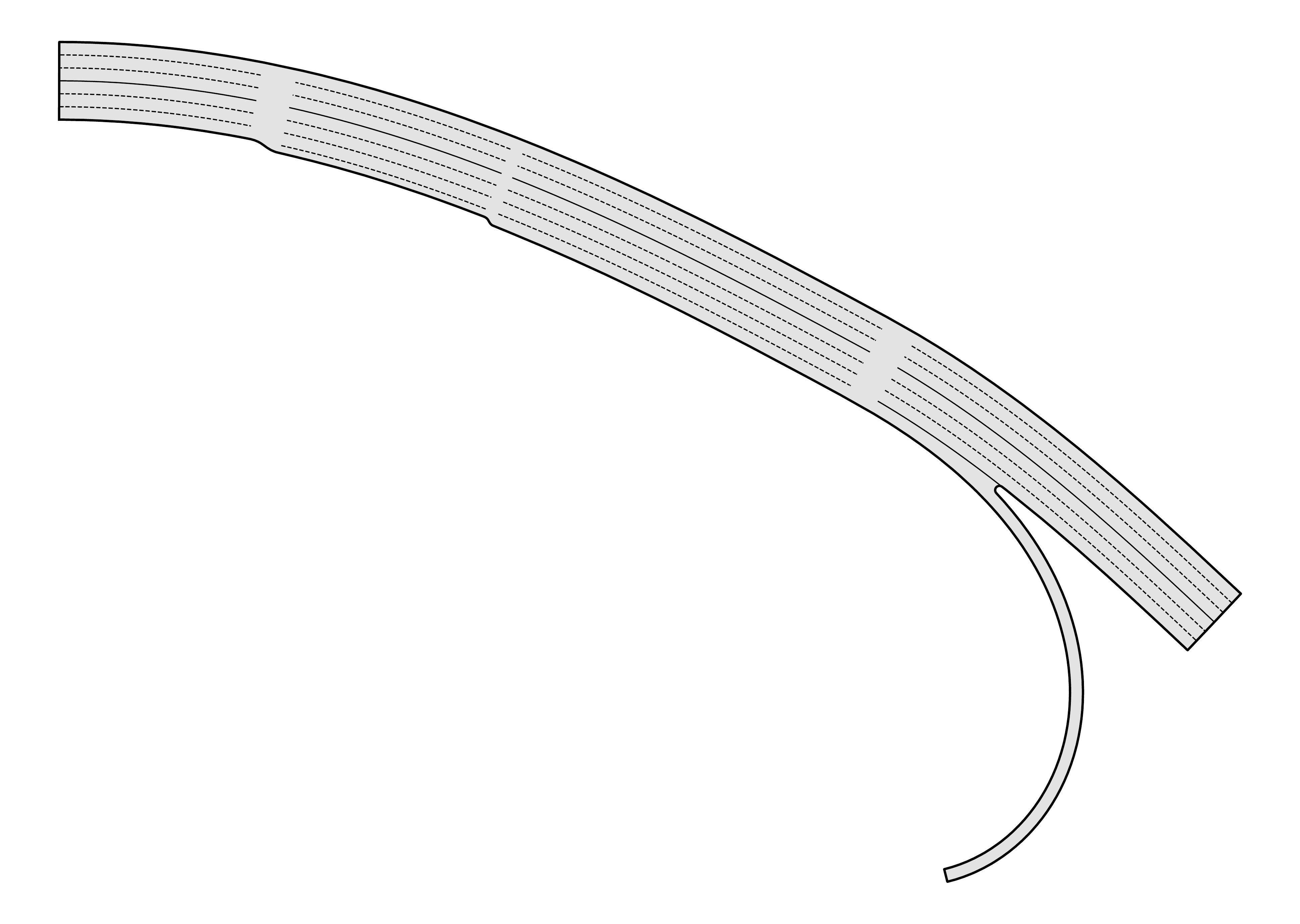}}

\subfigure{\includegraphics[width=.18\linewidth]{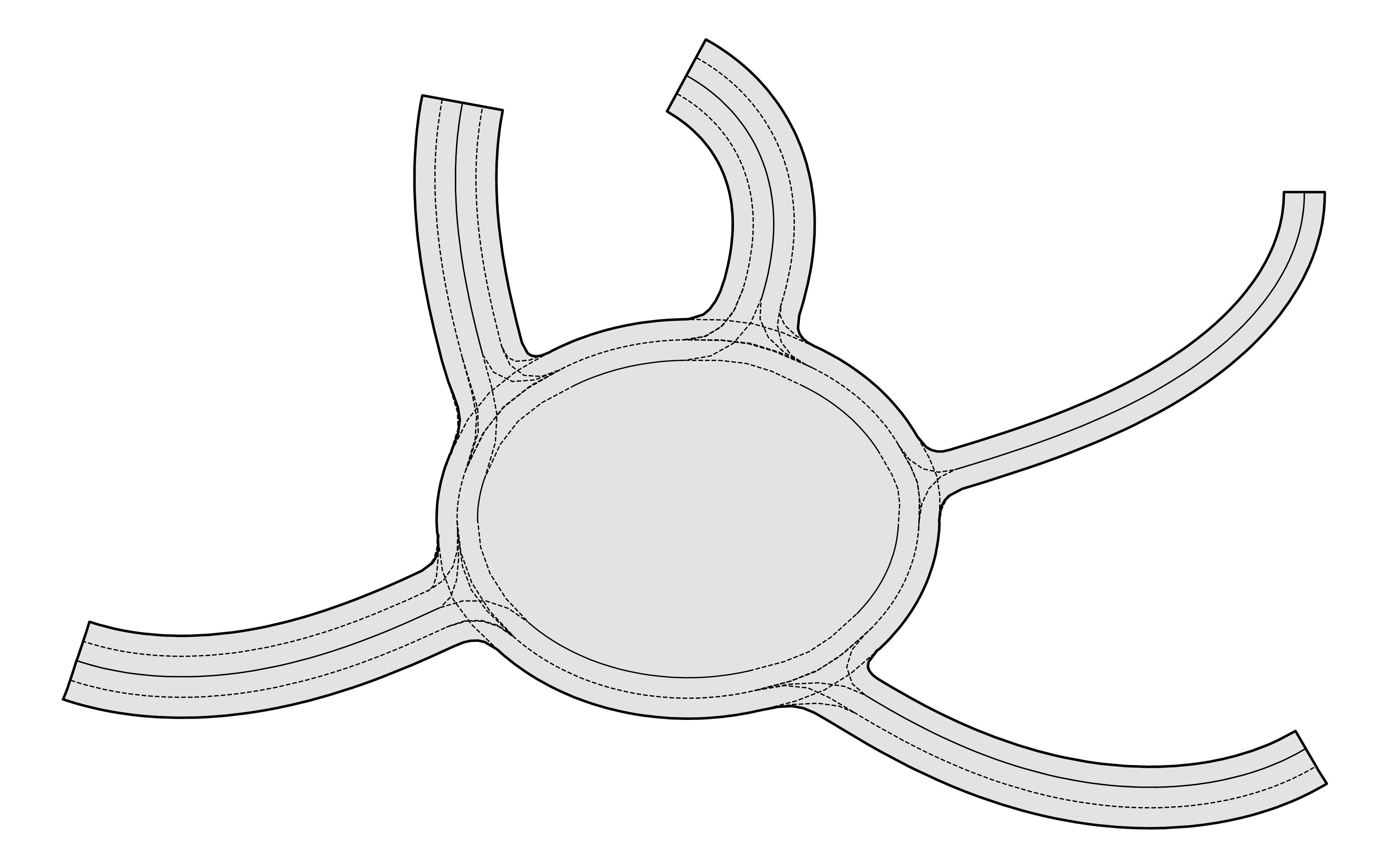}}\hfill
\subfigure{\includegraphics[width=.18\linewidth]{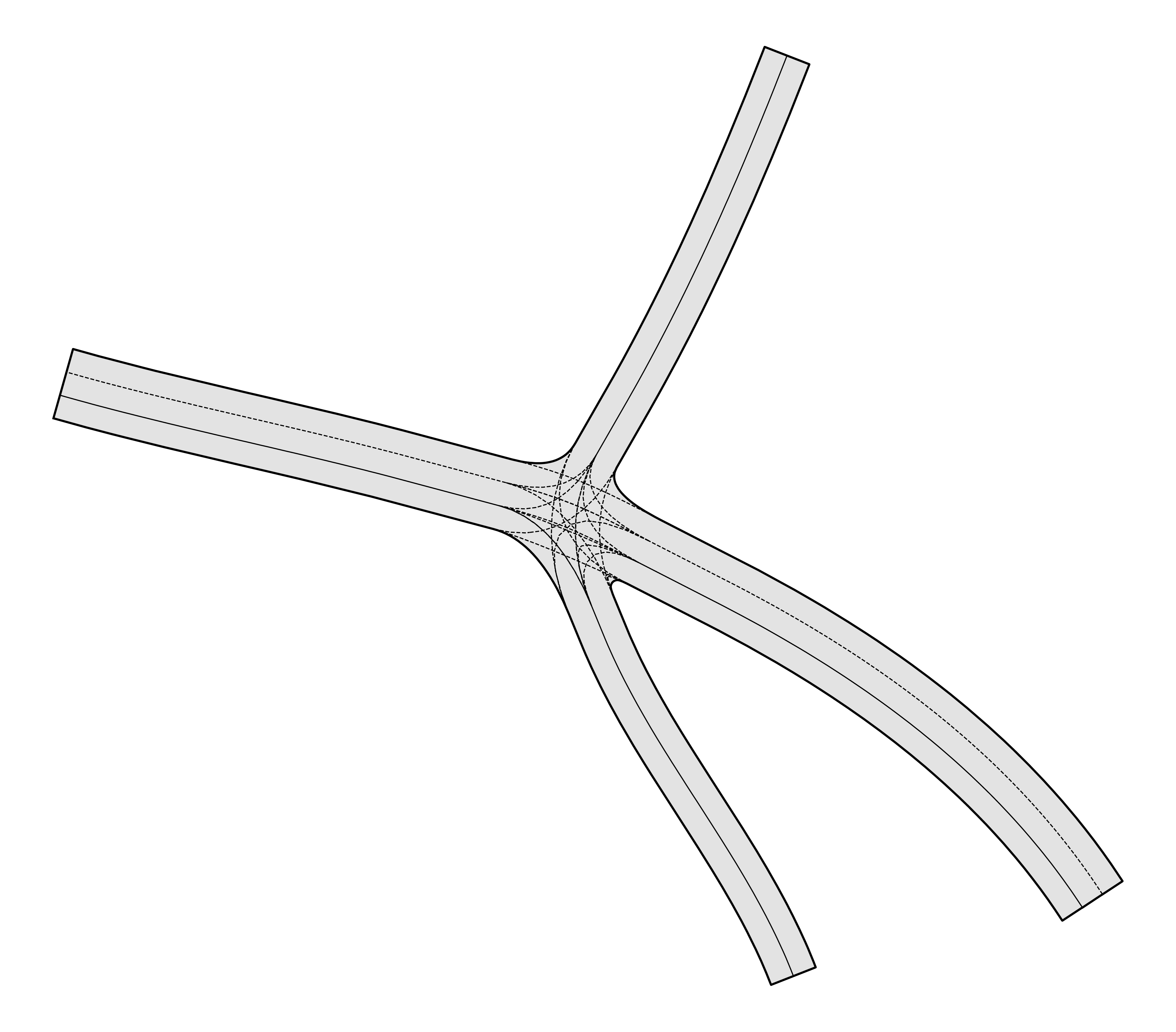}}\hfill
\subfigure{\includegraphics[width=.18\linewidth]{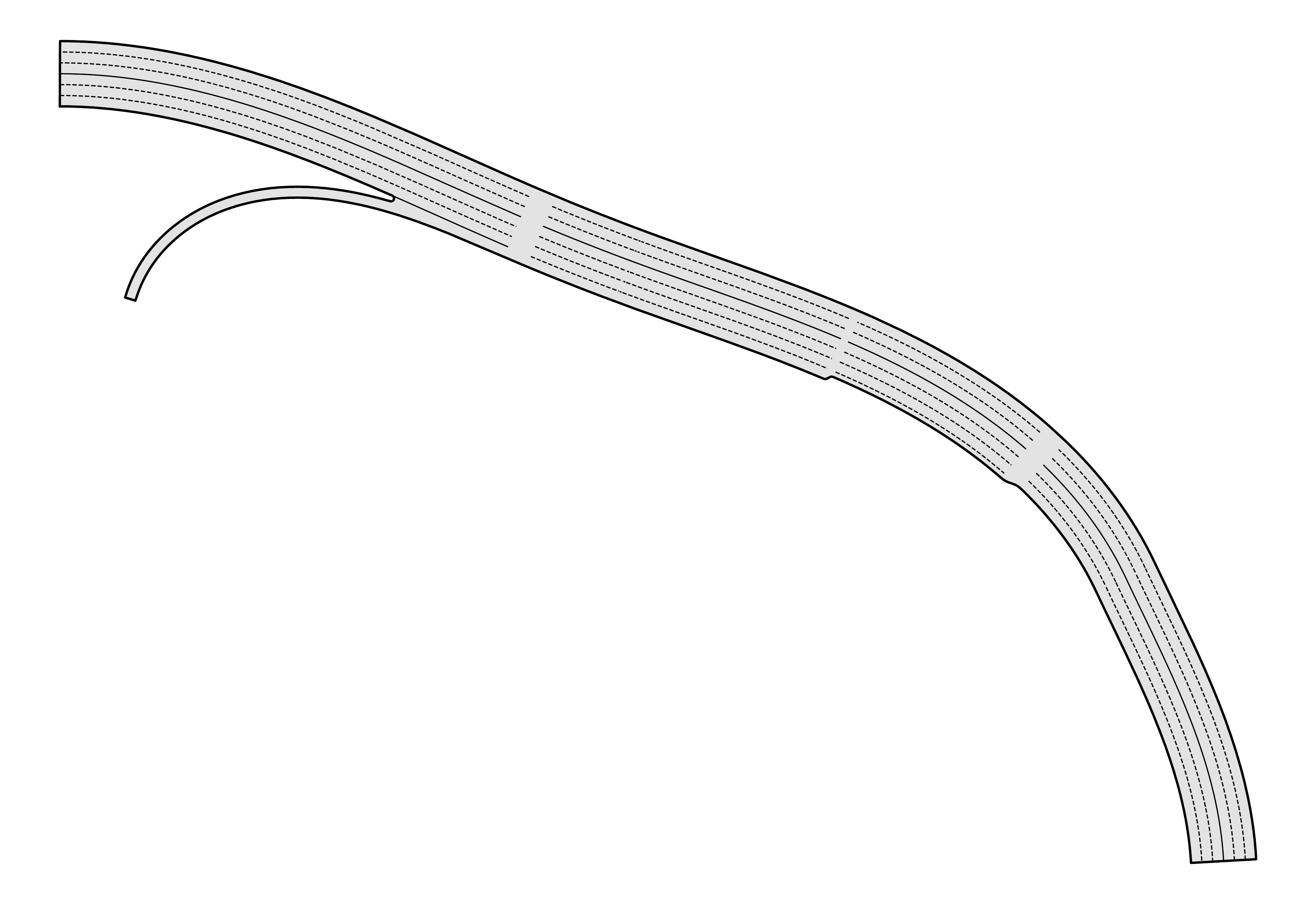}}\hfill
\subfigure{\includegraphics[angle=90,origin=c,width=.18\linewidth]{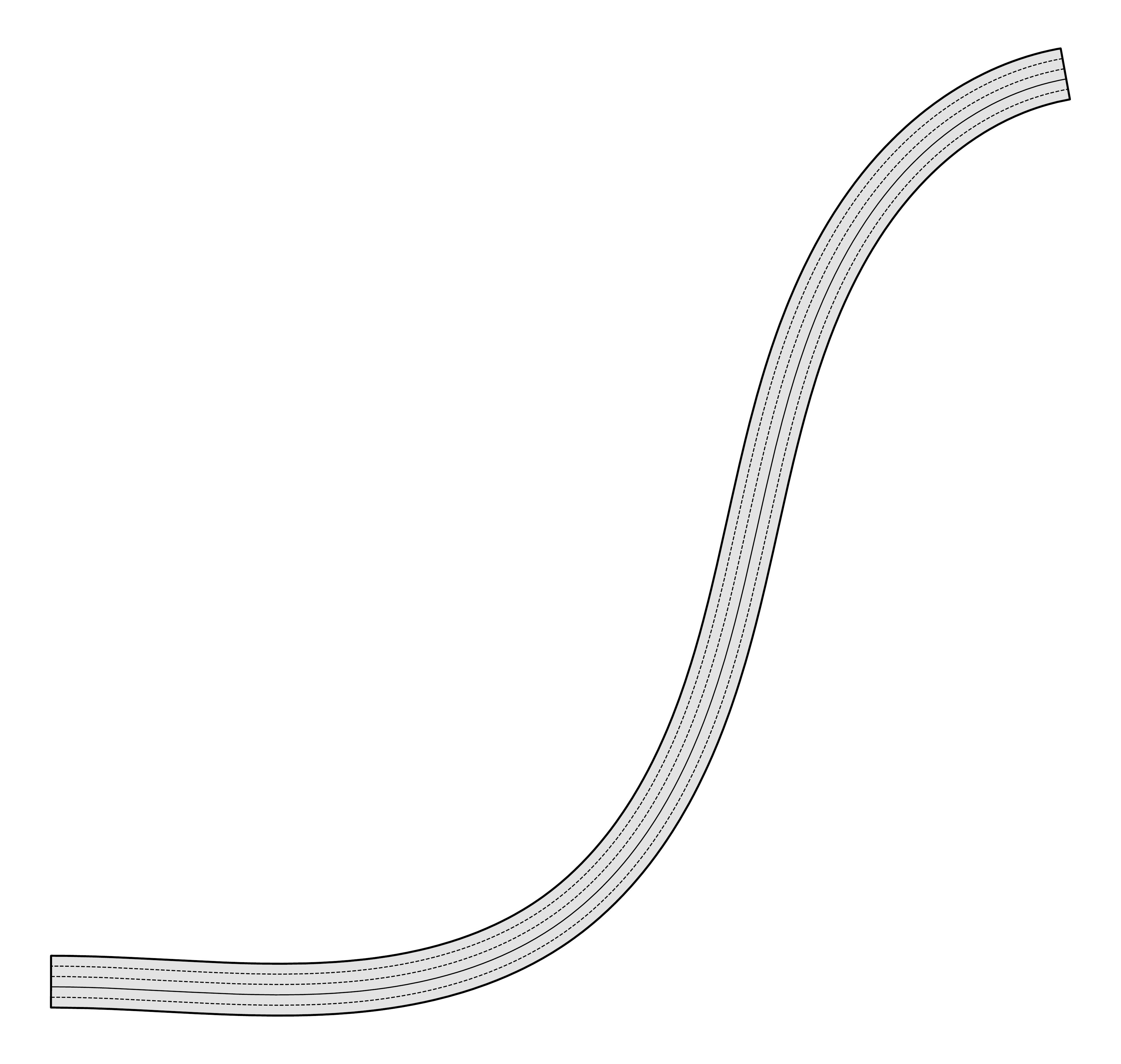}}\hfill
\subfigure{\includegraphics[width=.18\linewidth]{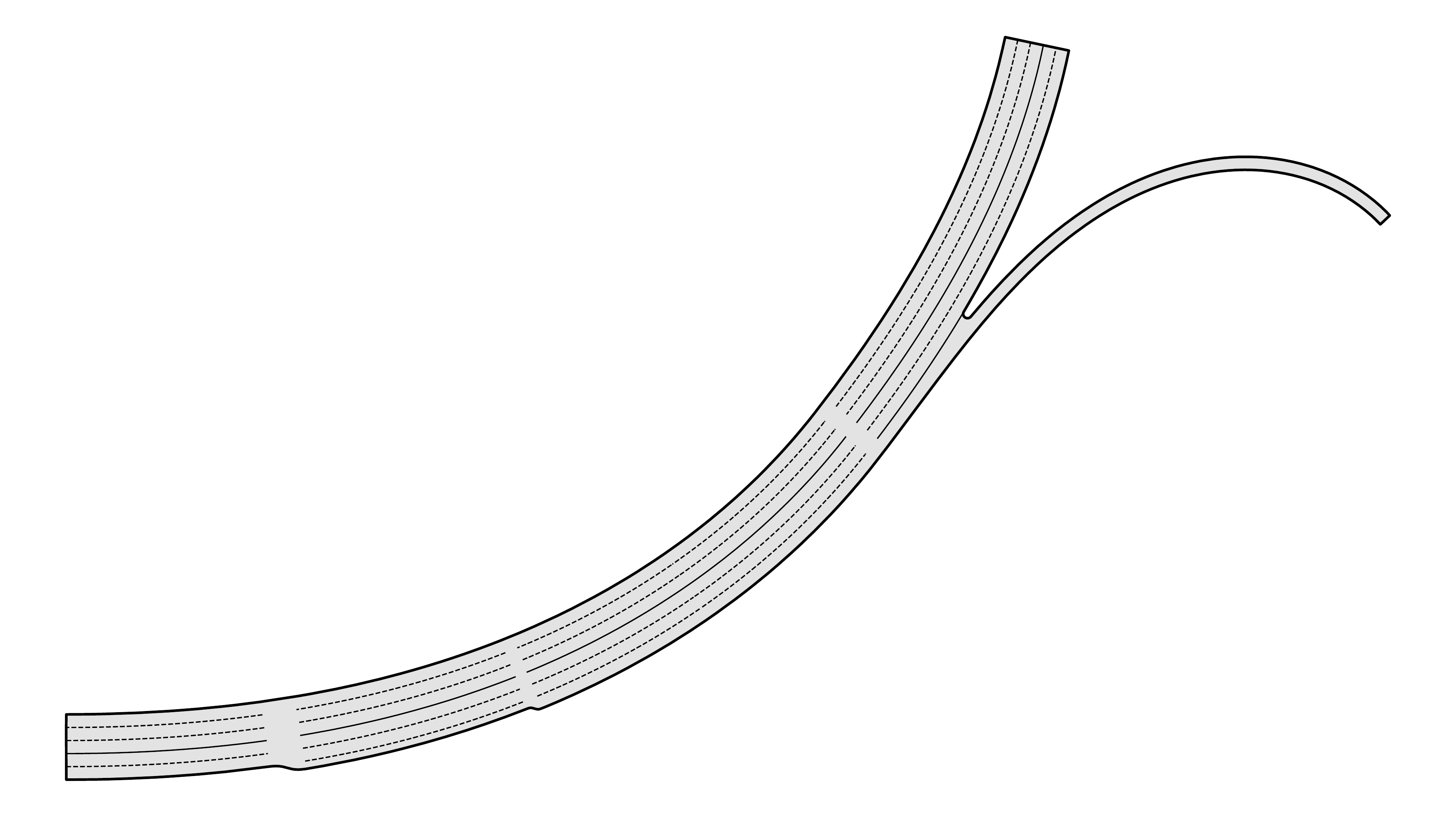}}

\caption{Generated maps for the different scenarios. For each of the five scenarios roundabout, intersections, highway entry, drive and exit (left to right), two examples are shown, one per row.}    
\label{fig:generatedmaps}
\end{figure}

\subsection{Randomization}
\label{subsection:dojo:5}

We randomize on both the street network level (Section~\ref{subsection:randomization:street}) and the traffic level (Section~\ref{subsection:randomization:traffic}) on multiple levels and through different methods. Our core benchmark scenarios include roundabouts, intersection, and highway-entry, highway-drive and highway-exit. Examples are shown in Figure~\ref{fig:generatedmaps}.

\subsubsection{Generation of Street Networks}
\label{subsection:randomization:street}

To offer maximum flexibility for creating highly diverse scenarios, Driver Dojo offers programmatic scenario definitions leveraging the \texttt{scenariogeneration}-package introduced in Section~\ref{subsection:oss}, through functions of the official \texttt{sumolib}-package. 
While the former builds on the OpenDRIVE standard and requires translation into the SUMO format during runtime, the latter allows for generating native network definitions. From our experience, however, this has no significant influence on the actual runtime performance.
We translate between conventional map formats, including OpenDRIVE, through netconvert, which is part of the SUMO software ecosystem. 
As an additional third option, we further integrate netgenerate into our benchmarking suite, which is a commandline tool that is especially powerful when targeting larger grid-like networks.

For our main benchmark scenarios, we use clothoids to approximate realistic road curvatures and define distributions over start and end curvatures and the number of right and left lanes of road segments to introduce variety.
For the roundabout and intersection scenarios, we define distributions over the number of incoming lanes, their angle in relation to the structural center, their lengths, as well as the distance of their connection point with the junction. For roundabouts in particular, we added probabilistic deformations in $x$- and $y$-direction, as real-world roundabouts are often not perfectly round. More details can be found in the supplementary material. 

\subsubsection{Traffic Modelling and Driving Personalities}
\label{subsection:randomization:traffic}

SUMO's microscopic traffic simulation modularizes traffic behavior through separate car-following, lane-change, and junction models. Commonly, such models are fixed for one traffic simulation, but different model parameters can be assigned to groups of vehicles.
These span across attributes that define, for instance, pushiness or willingness for strategic lane changes, but also safety-related factors like targeted time headway, driving imperfections or malicious overlooking when crossing a junction.
To make our benchmark as challenging as possible, we define distributions over 34 such parameters of interest, from which we sample a fixed set of parameter constellations ($200$ in the default case), and randomly assign them to non-egos entering the scenario. These also include physical properties such as vehicle dimensions or acceleration and deceleration profiles. More detailed information is given in the supplementary material. 

To make the environment as challenging as the real world, we introduce unexpected traffic events such as emergency break triggers for vehicles inside the cone in-front of the ego agent and softer speed variations inside a predefined radius.



\section{Experiments}
\label{section:experiments}

We tested popular RL algorithms on our Driver Dojo benchmark. 
To underline the necessity to compare AD methods on a common code base, we first study the impact of environment design choices on driving performance and generalizability in Section~\ref{subsection:design-choices}. 
Next, we present the results on the core scenarios of our benchmark in Section~\ref{subsection:eval-generalization}.
To test for generalizability, we train each algorithm on a fixed number of levels and evaluate on different set of levels. 
We use algorithm implementations offered by the Tianshou~\cite{weng2021tianshou} framework and train two agents on different seeds per model to account for stochastic variations in the training process. For all our experiments, we set the time resolution to $200ms$.
Performance is quantified as the Interquartile Mean (IQM) reward, which is the median performance of the middle $50\%$ of runs and promises to be more robust to outliers while being more statically efficient than median performance~\cite{DBLP:journals/corr/abs-2108-13264}.
We additionally report the mean crash rate (CrR) and mean completion rate (CoR) of evaluation runs.
Hyperparameter settings are listed in the supplementary material. 
Regarding computational complexity, training a PPO agent for 10M time-steps using eight parallel environments required us roughly 24 hours on an AMD Ryzen 9 5900X 12-core workstation with an RTX 3900, which is very fast and allowed for fast prototyping.

\begin{figure}[t!]
    \centering
    \subfigure[Highway-Entry]{\includegraphics[trim=35 0 10 0,width=0.32\textwidth]{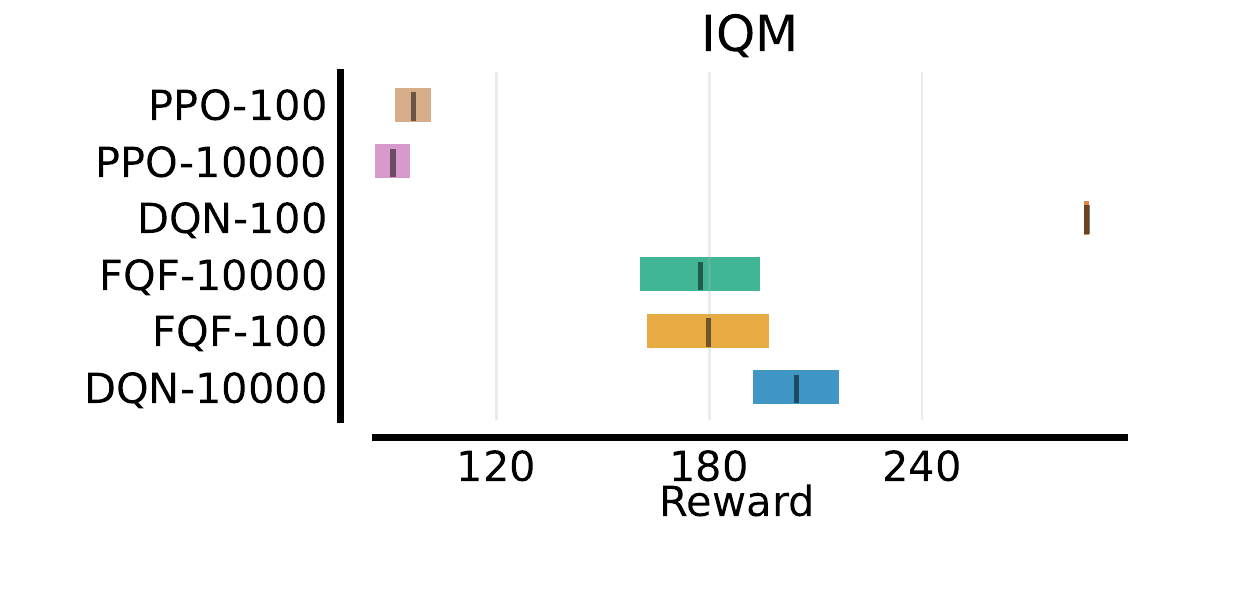}}
    \subfigure[Intersection.]{\includegraphics[trim=35 0 10 0,width=0.32\textwidth]{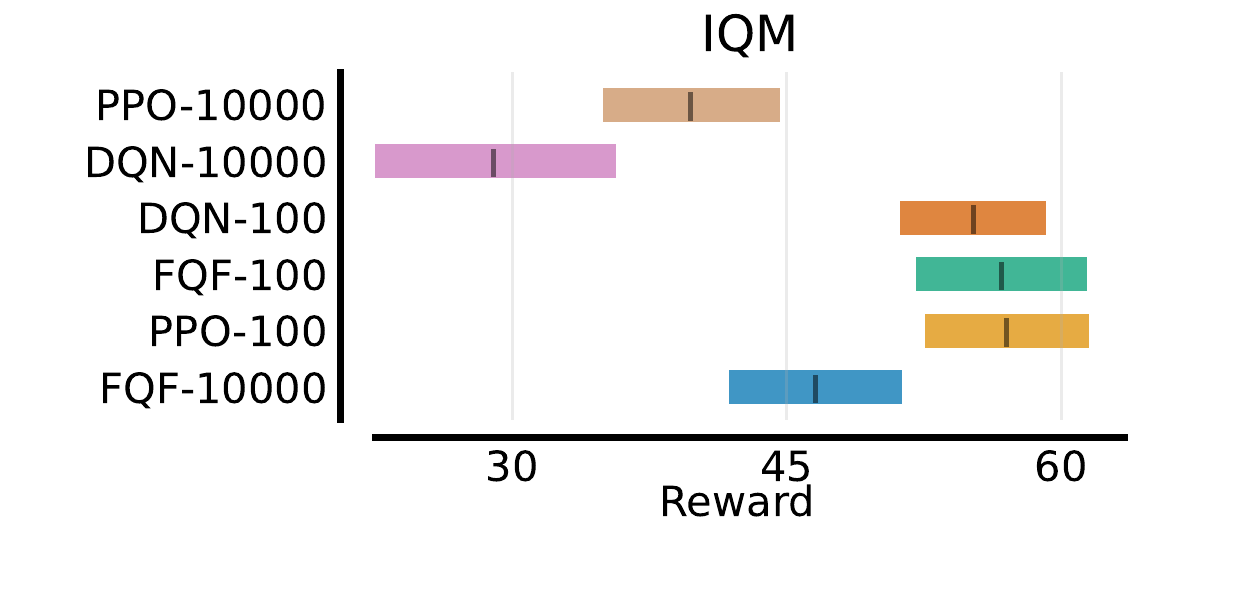}}
    \subfigure[Roundabout.]{\includegraphics[trim=35 0 10 0,width=0.32\textwidth]{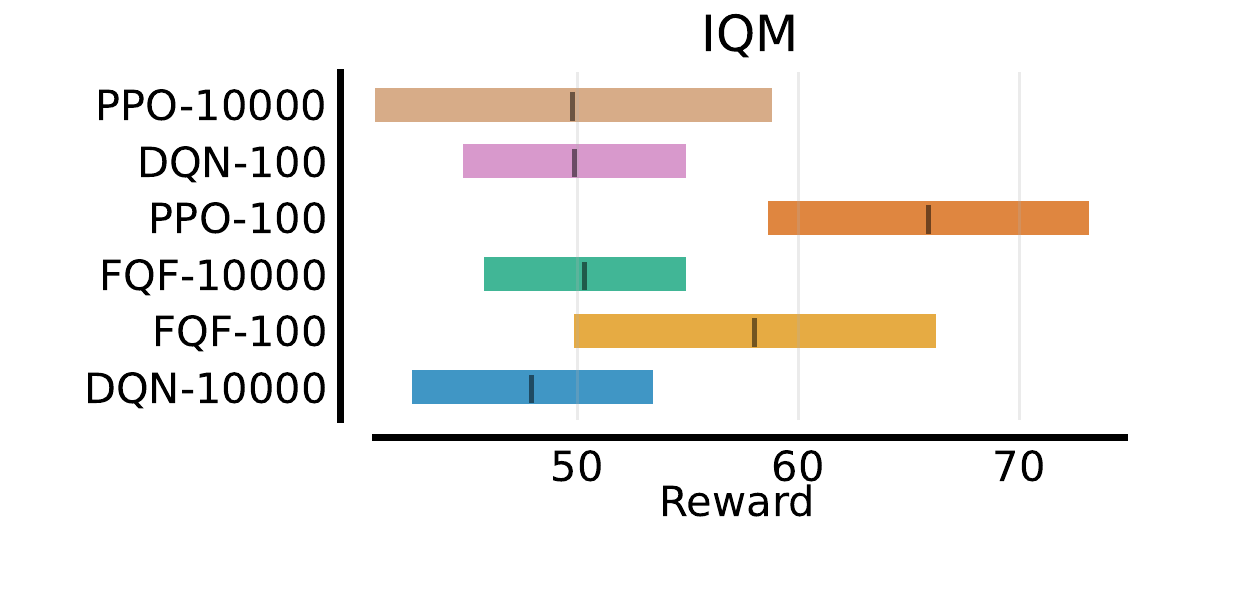}}\hfill
    \caption{Results from the generalization experiments on the different scenarios: PPO, DQN and FQF trained on either 100 or 10,000 different configurations and evaluated on 100 different configurations. }
    \label{fig:my_label}
\end{figure}

\subsection{Impact of General Design Choices}
\label{subsection:design-choices}

To efficiently compare different design choices, we fix the algorithm to PPO~\cite{schulman2017proximal} and vary specific aspects of the environment setup. Models were trained on 1K different levels.

We use a continuous action space (PPO-Cont.), its variate where steering velocity and pedal inputs are discretized into sets of five equidistant values (PPO-Disc.), as well as a semantic action space (PPO-Sem.).
In general, we let the agent perceives the environment as a feature-space spanned by the combination of EgoState, TrafficState, RoadShape and Navigation observers. 
Our fourth agent receives image-based BirdEye observer outputs instead (PPO-Sem-BEO). To ensure the Markov property, we use frame-stacking of five.
For direct actions we used the kinematic single-track dynamics model, for semantic actions our special TPS vehicle.
In general, we use the reward function:

\begin{equation}
    R_a(s_t, s_{t+1}) = 
    \begin{cases}
        5.0 & \text{if $s_{t+1}$ is a sub-goal state} \\ 
        10.0 & \text{if $s_{t+1}$ is a goal state} \\
        -10.0 & \text{if $s_{t+1}$ is a crash, off-route or non-road state} \\
        v_{t+1}/v_{max} & \text{otherwise, where } v \text{ is the velocity.}\\
    \end{cases}
\end{equation}

It has multiple sparse reward components and a dense speed-reward to guide the agent towards a more efficient driving behavior.

\begin{wrapfigure}{r}{7cm}
    \centering
    \vspace{-5mm}
    \includegraphics[width=0.4\textwidth]{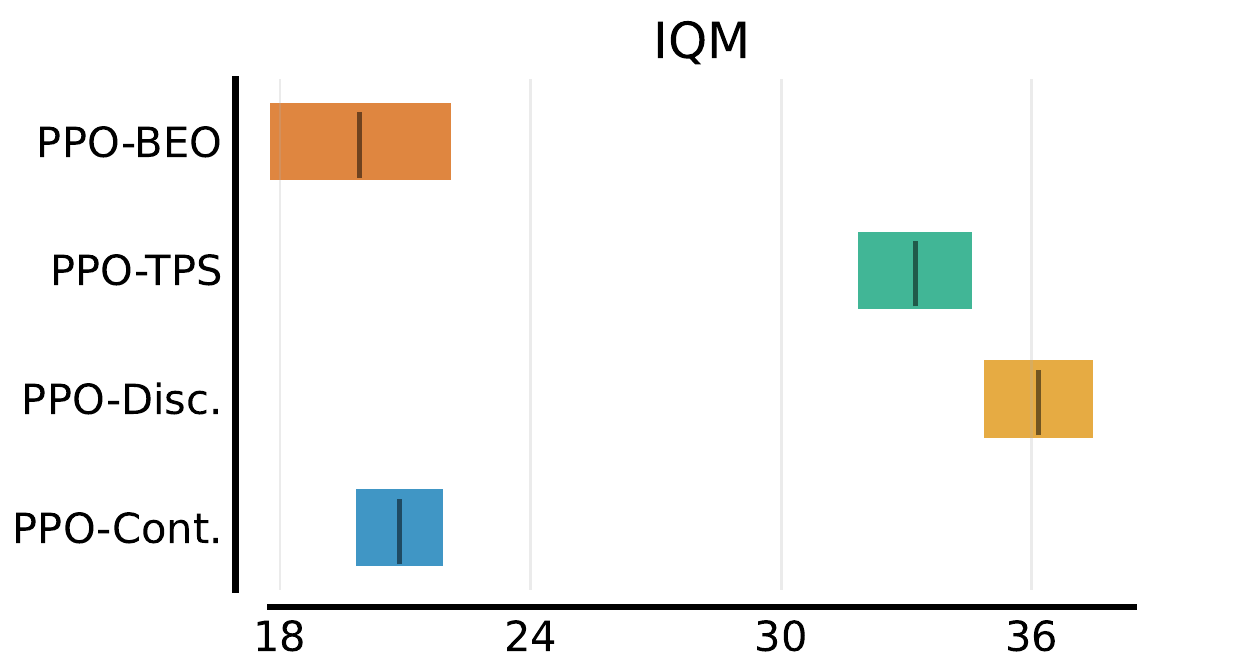}
    \caption{IQM performance of PPO agents trained on the same sets of 1M training scenarios.}
    \vspace{-5mm}
    \label{fig:design-choices}
\end{wrapfigure}
\textbf{Results.} The results of our experiments are shown in Figure~\ref{fig:design-choices}. We see that small changes in the environment design can have a non-negligible impacts on learning performance. Here, the discrete action space wins the race, closely followed by the semantic-actions agent. The same agent with the BirdEye observer attached does not perform well, which might be due to the fact that visual models tend to require a lot training time due to the complex nature of reward-based feature learning through CNNs. PPO-Continuous achieved similar performance as the CNN-based agent, which hints to the fact that continuous action spaces might be harder to generalize from in an AD setting.

\subsection{Benchmarking for Generalization}
\label{subsection:eval-generalization}

For our generalization benchmark, we aimed to cover popular algorithmic families and trained Double Duelling DQN~\cite{wang2016dueling} for off-policy RL, the distributional action-value based variant FQF~\cite{yang2019fully}, and the on-policy PPO as in the previous experiment. To underline the impact of variations in the training data on generalization to unseen scenarios, we used either 100 or 10K different training levels and evaluate them on the same set of test levels. Here, we limited ourselves to the semantic action space as we see it as the most promising approach to successfully teach RL agents to drive. In contrast to traditional training regimes, we collected experience for DQN and FQF in multiple parallel environments to ensure better comparability with PPO, where we used eight parallel training environments in total. A detailed description on our hyperparameter configuration can be found in the supplementary material.

\begin{table}[b!]
\caption{Crash rate (CrR) and completion rate (CoR) for agents trained on the Roundabout scenario using the TPS action space. Runs that neither completed nor crashed either exceed the environment time limit or ended with the agent driving off-route.}
\label{table:experiment2_rates}
\centering
\vspace{4pt}
\begin{tabular}{lll}
\toprule
Agent & CrR & CoR \\
\midrule
DQN-100 & 31.07 & 38.00 \\
DQN-10000 & 33.71 & 37.60 \\
PPO-100 & 35.70 & 43.63 \\
PPO-10000 & 33.71 & 37.68 \\
FQF-100 & 21.81 & 34.04 \\
FQF-10000 & 23.80 & 43.05 \\
\bottomrule
\end{tabular}
\end{table}

\textbf{Results.} The results of our experiments are illustrated in Figure~\ref{fig:my_label}. The numbers indicate the number of training environments, e.g., PPO-100 means training took place on 100 different environments. We tested every agent on a set of 2K unseen scenarios in total. Surprisingly, we observe that the evaluation performance is generally much better for agents that have been trained on 100 scenarios compared to the agents that have been trained on 10K scenarios. 
This is in stark contrast to contributions like ProcGen~\cite{DBLP:journals/corr/abs-1912-01588}, where the number of training levels and test set performance tend grow proportionally with each other. We believe that a possible reason for this might be the network capacity of the actor and/or value network. A large distribution of different environments may result in noise-like gradient updated that lead to instabilities during training.

Methods that could be especially beneficial in our environment might be level replay and design methods such as PLR~\cite{jiang2021prioritized}, Repaired~\cite{dennis2020emergent} or ACCEL~\cite{parker2022evolving}. We want to note that it is straightforward to expose our street network and traffic initialization scheme through an action space suitable for level-design but we leave this for future work.

\section{Conclusion}
\label{section:conclusion}

We introduced Driver Dojo as a benchmark for prototyping, training and evaluating agents across wide ranges of scenario variations. It allows to generate a wide range of different driving scenarios under extended randomization. Apart from pre-implemented scenarios, we elaborated how Driver Dojo allows for fast and efficient prototyping of new training environments and solutions. In the experiments we showed that, compared to other many other benchmark, the scenarios offered by Driver Dojo are hard to solve and proper algorithms are needed. Further, unsupervised level design methods becoming popular nowadays require special action spaces in order to dynamically adapt scenarios through outside control.

\textbf{Limitations}
Future work might include additional road scenarios and even more elaborate traffic modeling, as we only cover intersection, roundabouts and highways with our core benchmark. At the moment, Driver Dojo is also limited to traffic made up of passenger cars solely, which still represent a large gap to the real world. Further, static and moving obstacles like pedestrians are still missing in our benchmark. This goes further to obstacles that obscure the view on places like intersections, where real-world drivers would be required to advance very slowly.

\textbf{Possible negative impacts.}
The purpose of the benchmark is to allow for development of generalizable and safe autonomous driving algorithms. Autonomous driving in general is not deemed to have large negative effects on society. Current developments in autonomous driving aim more at assisting human drivers and thereby reducing accidents and increasing personal freedom. Certain jobs like taxi drivers might be less in demand in the longer term. These developments are not substantially influenced by our work, as we only foster the testing for safety and dependability of algorithms independently developed from our benchmark.

\section*{Acknowledgements}

This work was supported by the Bavarian Ministry for Economic Affairs, Infrastructure, Transport and Technology through the Center for Analytics-Data-Applications (ADA-Center) within the framework of “BAYERN DIGITAL II”. 

\clearpage
\appendix
\section{Appendix}

This supplement provides additional information and documentation for Driver Dojo and the settings that we have used for the experimental section of the paper. Please find an overview of the available sections below:

(\ref{sec:simparams}) Randomized Simulation Parameters\\
(\ref{sec:setup}) Experimental Setup and Hyperparameters\\
(\ref{sec:results}) Supplementary Results

\subsection{Randomized Simulation Parameters}
\label{sec:simparams}

In this section, we provide more intricate details on different randomization mechanisms used in our benchmark. This includes traffic and ego initialization and routing, parameter distributions used to inject independent driving personalities into the simulation, and the way core scenario maps are generated. First, we give details about how we seed randomness in Driver Dojo.

\subsubsection{Seeding} 

To allow for full reproducability, we decouple the random number generation process through multiple, independent number generators. The overall environment seed is used to seed a master generator, which, after every environment reset, produces a new seed for a random number generator object used for map generation, and a second seed used for the traffic scenario generator. This way, no matter how long and in which way the agent interacts with the environment, it is always guaranteed that, based on the same initial master seed, the same succession of traffic constellation and behavior and street networks is generated. The master generator is exclusively used to create the map and traffic random generator seeds.

Further, this seeding process allows us to constraint the number of maps, as well as the number of traffic situations cycled through during the whole training duration. In its current state, however, the seeds created to initialize the constrained set of scenarios depend on both constraint values. This means that, for example, the setup where we visit 100 different maps with 1 traffic scenario each contains different maps and traffic situations than a setup where we visit 100 maps with 2 traffic scenarios each. This, however, should only pose as a disadvantage in very specific experimental settings and can also very easily be improved upon.

\subsubsection{Traffic and Ego Initialization and Routing}
At the beginning of an episode, we seed every scenario with initial non-ego traffic based on a density metric. In our case, we spawn vehicles on every road that is not a connecting road inside a junction, such that vehicles have an approximate distance of $30$ meters between each other. This only holds approximately true, as we let the SUMO engine handle the exact details of traffic participant insertion and let vehicles be spawned on the "best" lane, where best means one of the lanes from which a vehicle has to perform a minimum amount of lane changes to arrive at its destination. We observed less unwanted congestions inside the scenario using this method. Based on the insertion road, every vehicle is routed to a random, reachable exit point from the map.

We spawn the ego-vehicle at a random free position into the scenario. In the Roundabout and Intersection scenario, this is one of the incoming lanes, for Highway-Entry scenario the lanes on the entry road, and for the other two scenarios a lanes on the first road of the highway section. For further variety we also randomized the initial velocity at spawn time.

\subsubsection{Traffic Model Parameter Sampling for Driving Personalities }

\begin{figure}
    \centering
    \subfigure[General]{
        \includegraphics[width=0.8\textwidth]{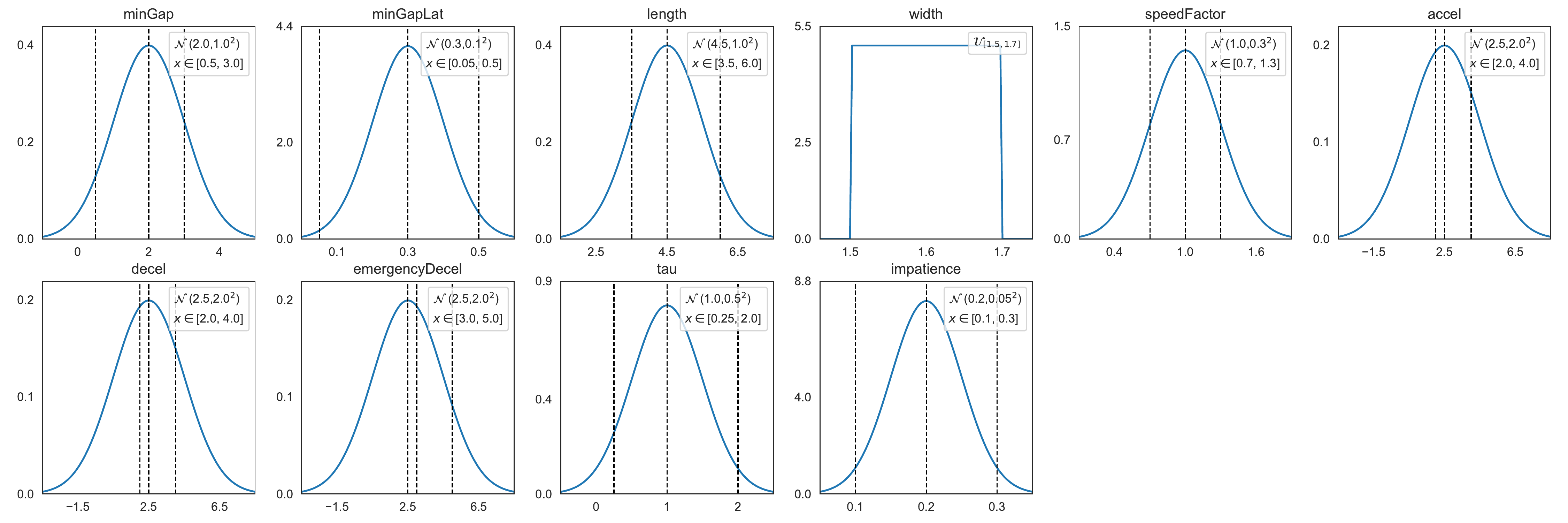}
        \label{subfig:vtype-general}
    }
    \subfigure[Car-following Model]{
        \includegraphics[width=0.8\textwidth]{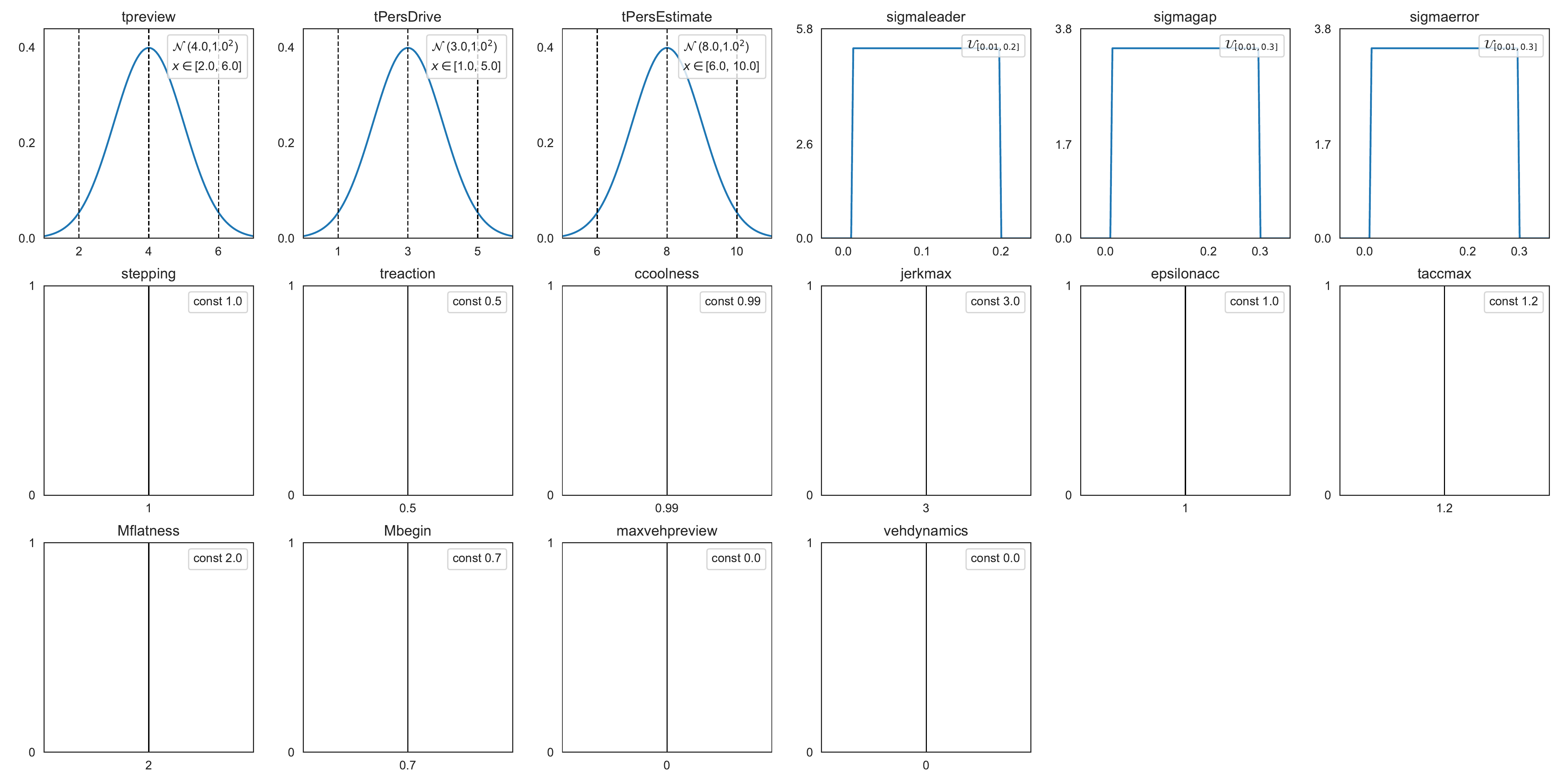}
        \label{subfig:vtype-carfollow}
    }
    \subfigure[Lane-change Model]{
        \includegraphics[width=0.8\textwidth]{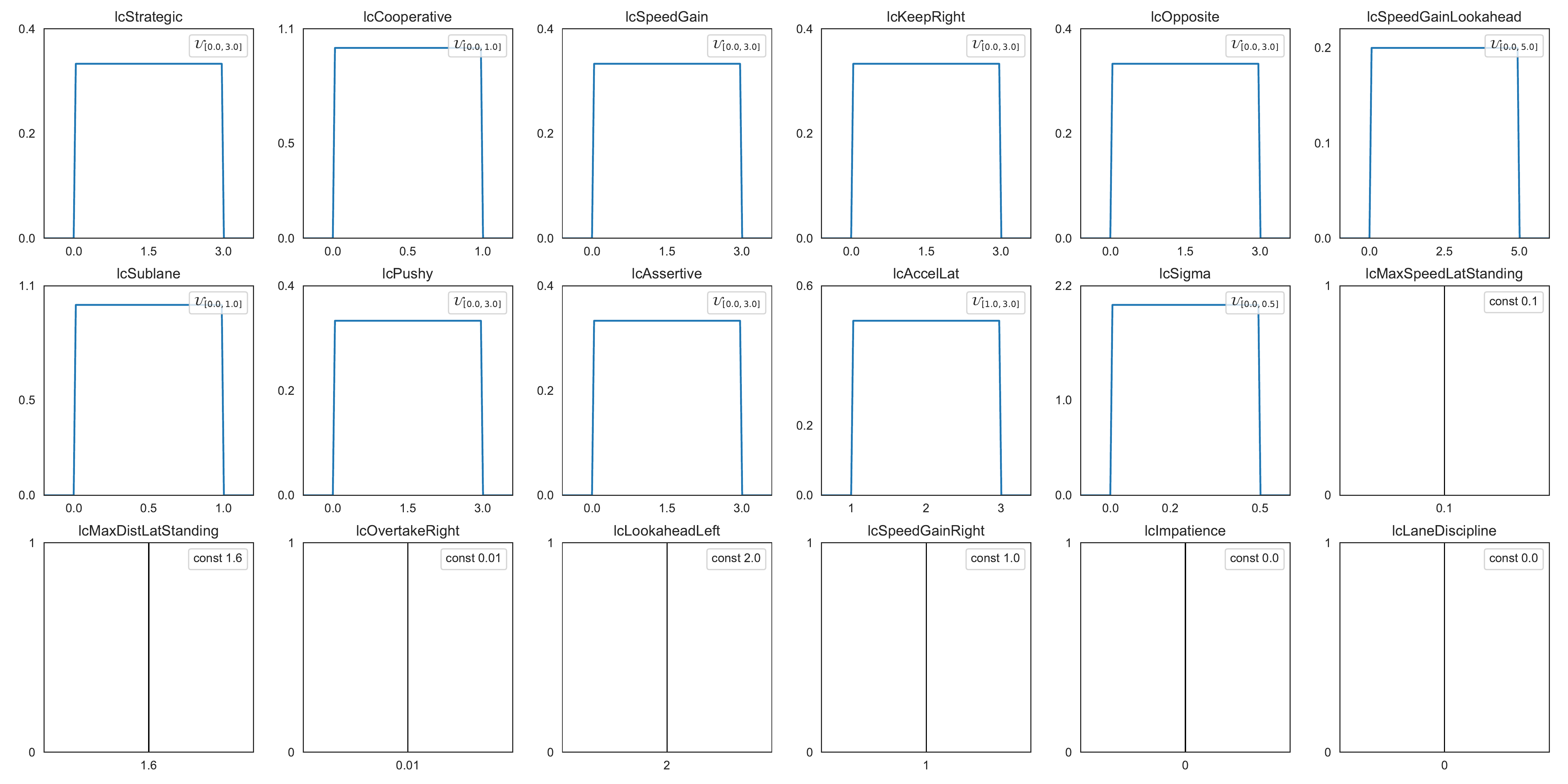}
        \label{subfig:vtype-lanechange}
    }
    \subfigure[Junction Model]{
        \includegraphics[width=0.8\textwidth]{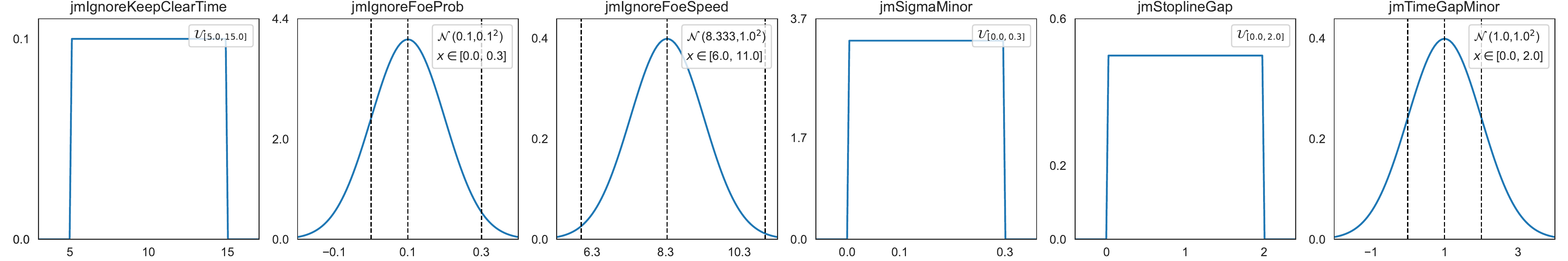}
        \label{subfig:vtype-junction}
    }
    \caption{SUMO vType sampling distributions and constants used to inject different driving personalities into the simulation. Inside the code base, these are defined in a separate text file, which allows for full configurability. For parameters drawn from a Gaussian distribution, we further clip samples into predefined ranges, indicated through doted vertical lines. }
    \label{fig:vType-dist}
\end{figure}

As was described in the main paper, we use randomly sampled general, car-following, lane-change and junction model parameters for non-ego traffic participants added to the simulated scenario. In more detail, we use SUMO's \texttt{createVehTypeDistribution} tooling script and create 200 a priori parameter constellations, which we randomly assign to every traffic member spawned. Figure~\ref{fig:vType-dist} shows the distributions and constants used in the default settings of our benchmark. For categorical parameters we used constant values as shown in Table~\ref{table:vType-dist}. Further, SUMO offers a range of car-following models to choose from, where we decided for the most recent EIDM~\cite{Salles2020ExtendingTI} model, which is the extended version of the famous IDM~\cite{idm2000} car-following model including many improvements from different contributions. It should further be noted that for urban scenarios we used a speed limit of $~13.889$ \si{m/s}, whereas in the highway scenarios we used $36.111$ \si{m/s}. In both cases, we constrain the maximum velocity of the ego vehicle to these values.

To give an intuition about behavioral aspects these parameters control, we give a rough overview about the effects of the most important ones.\footnote{The SUMO documentation provides a thorough explanation of every parameters, see \url{https://sumo.dlr.de/docs/Definition_of_Vehicles\%2C_Vehicle_Types\%2C_and_Routes.html}.} From the general category, besides self-explanatory attributes like \textit{accel} and \textit{decel}, \textit{speedFactor} defines a multiplicative factor causing deviations from the actual speed limit. On the other hand, \textit{tau} modifies the error proneness of the car-following model and \textit{impatience} influences the fundamental willingness to engage in dangerous driving maneuvers of the traffic participant, which further increases during waiting periods. For car-following parameters, \textit{t*} attributes define look-ahead time intervals and \textit{sigma*} define error factors, where the latter are exclusively applicable in case of the EIDM model. Lane-change model parameters predominantly modify the readiness to perform lane-changes w.r.t different objectives. As an example, \textit{lcSpeedGain} influences the inclination to change lanes and overtake a leading vehile to gain speed. Concerning the junction model the most interesting parameters, at least for our goals, are \textit{jmIgnoreFoeProb} and \textit{jmIgnoreFoeSpeed}, where the former defines the probability to overlook a junction foe and the latter the maximum speed of potential foes for such event to occur. Lastly, \textit{jmIgnoreKeepClearTime} incentives non-ego agents to drive over the stop line and wait inside the junction, even though passing the junction is currently not possible, a behavior many real-world drivers have internalized.

\begin{table}
\caption{vType constants used for categorical parameters for every traffic participant.}
\label{table:vType-dist}
\centering
\begin{tabular}{ll}
\toprule
Parameter & Value \\
\midrule
vClass & passenger \\
departLane & best \\
departPos & base \\
departSpeed & random \\
latAlignment & arbitrary \\
\bottomrule
\end{tabular}
\end{table}

\subsubsection{Map Generation}
\begin{figure}
    \centering
    \subfigure[Intersection]{
        \includegraphics[width=0.2\textwidth]{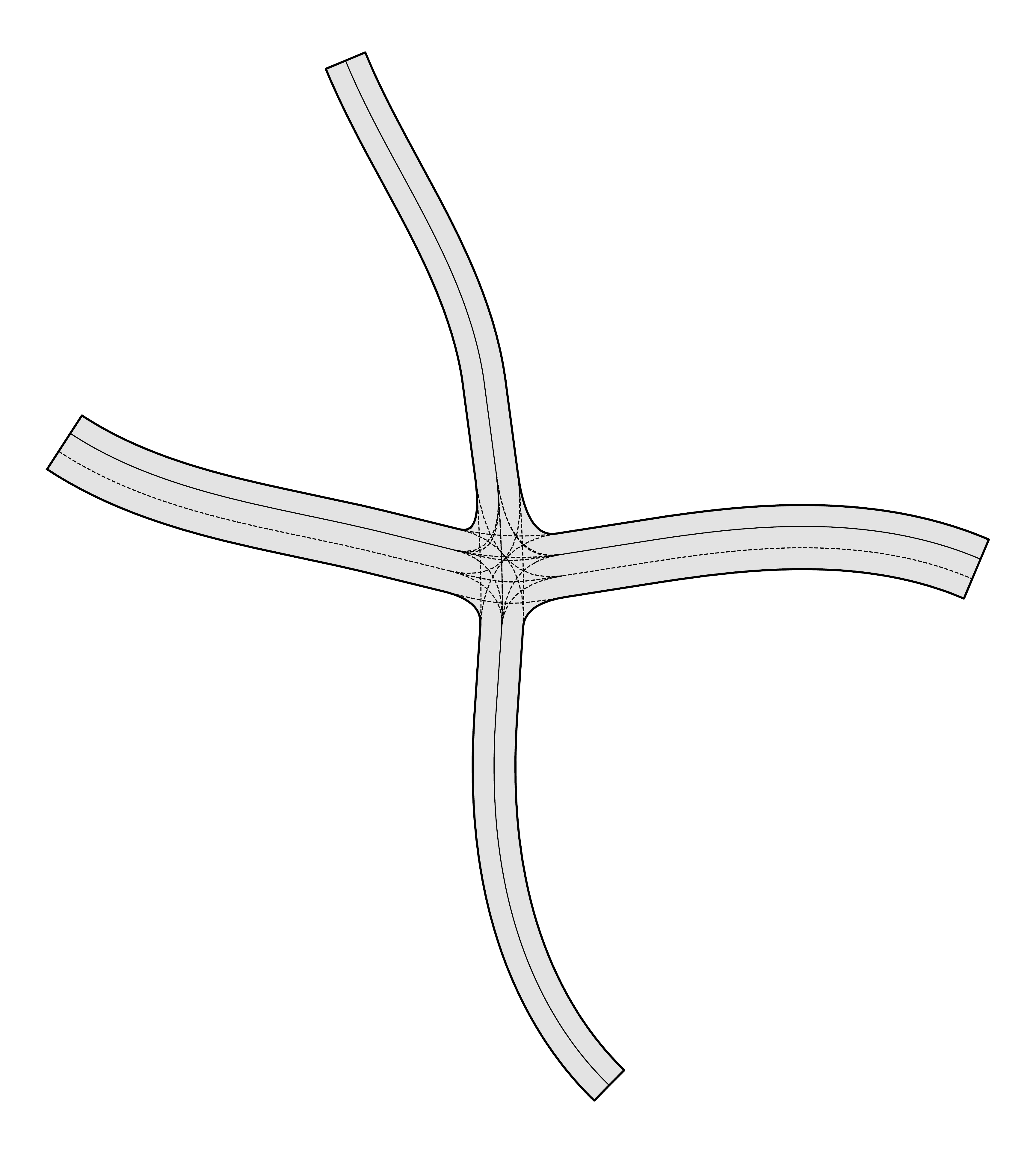}
        \includegraphics[width=0.2\textwidth]{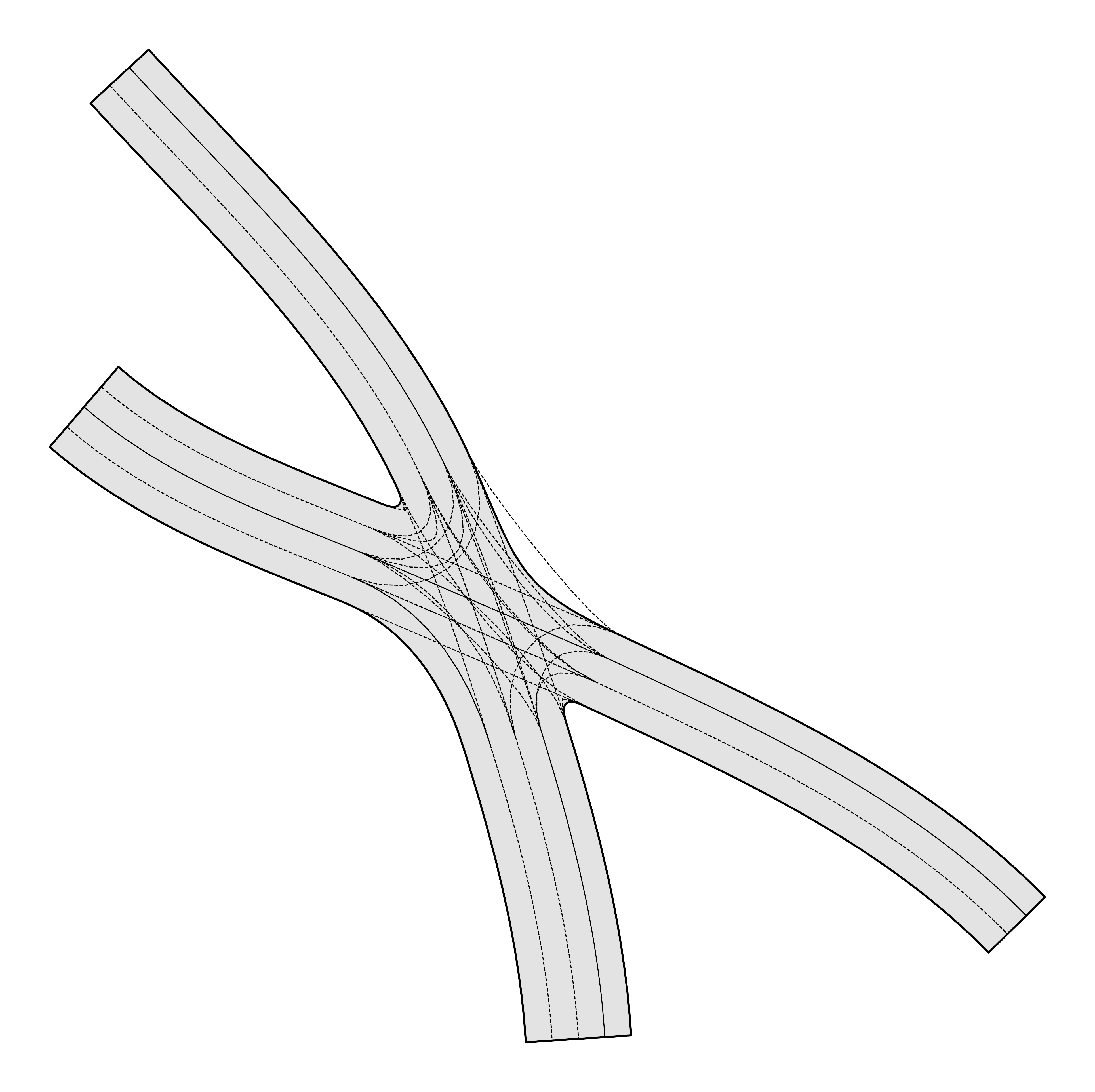}
        \includegraphics[width=0.2\textwidth]{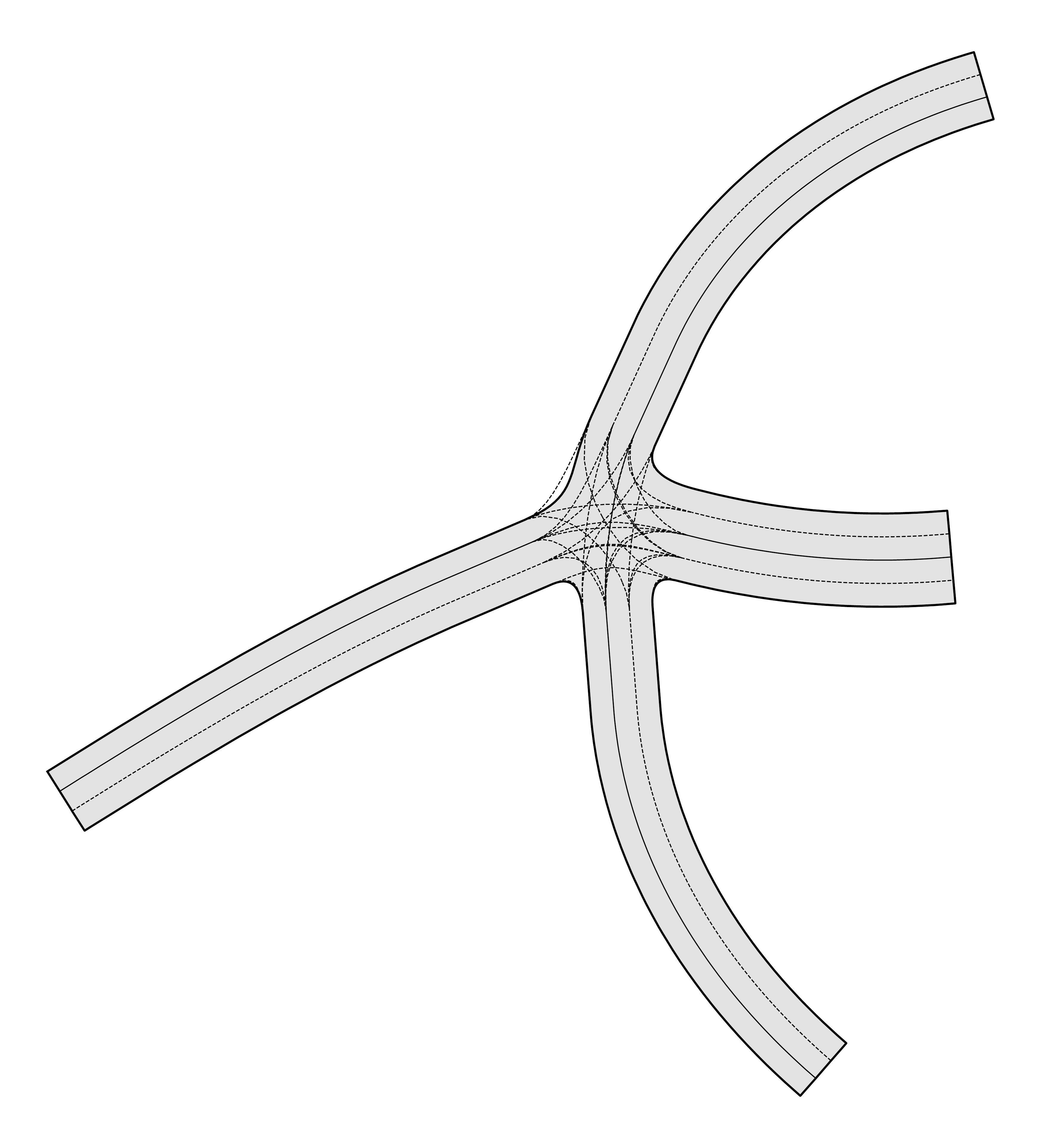}
        \includegraphics[width=0.2\textwidth]{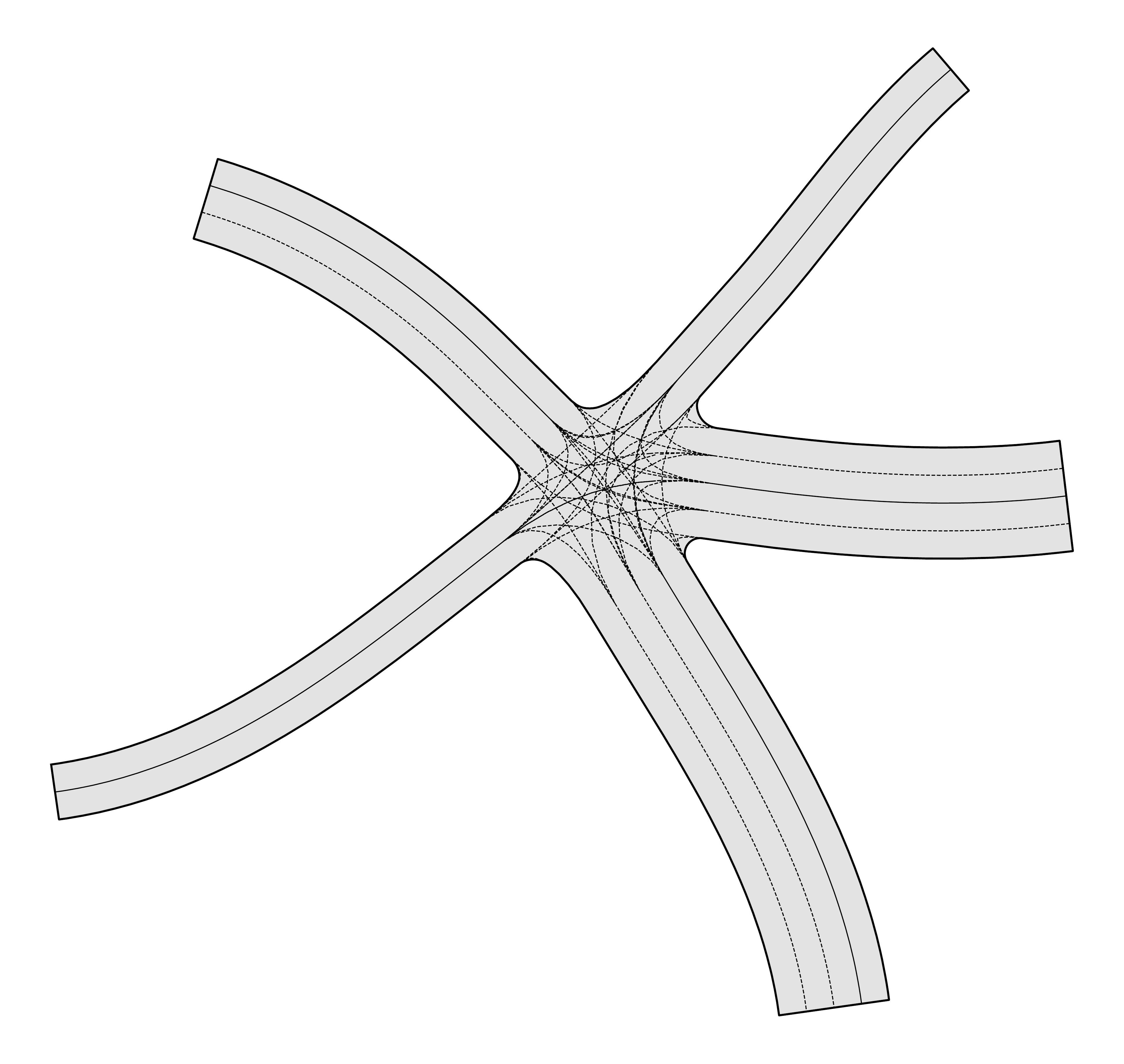}
        \includegraphics[width=0.2\textwidth]{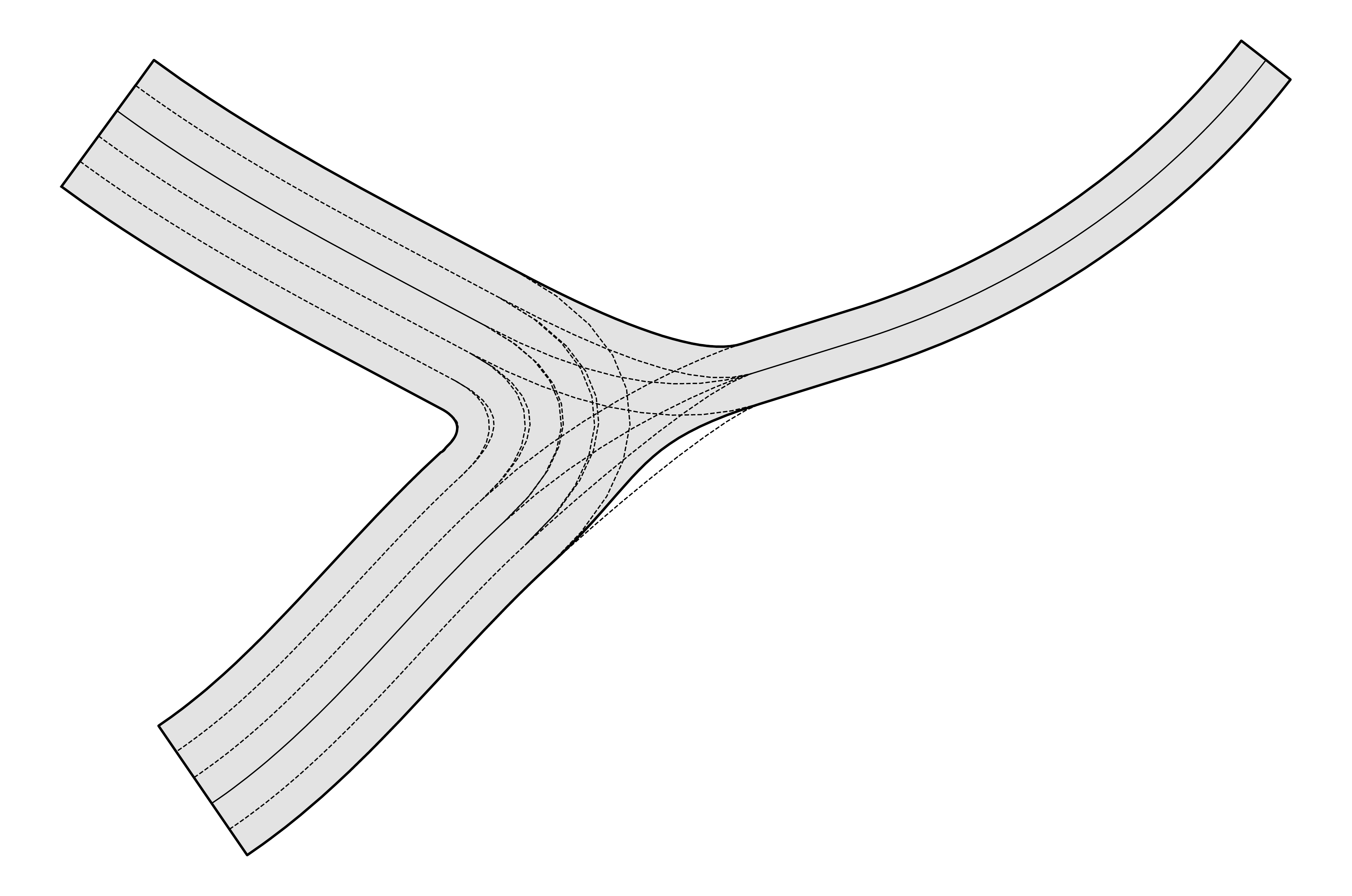}
    \label{subfig:map-vis-intersection}
    }
    
    \subfigure[Roundabout]{
        \includegraphics[width=0.2\textwidth]{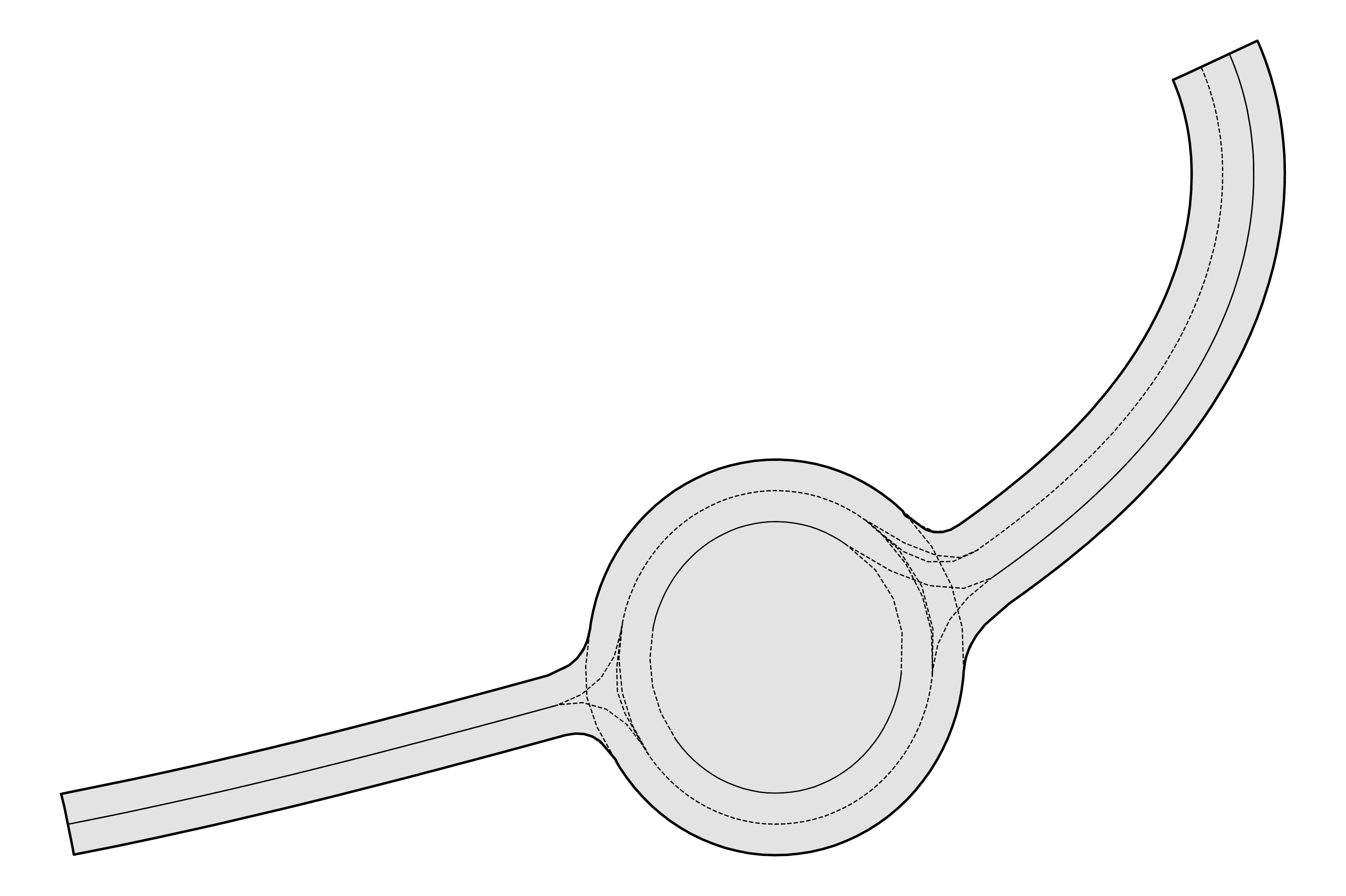}
        \includegraphics[width=0.25\textwidth]{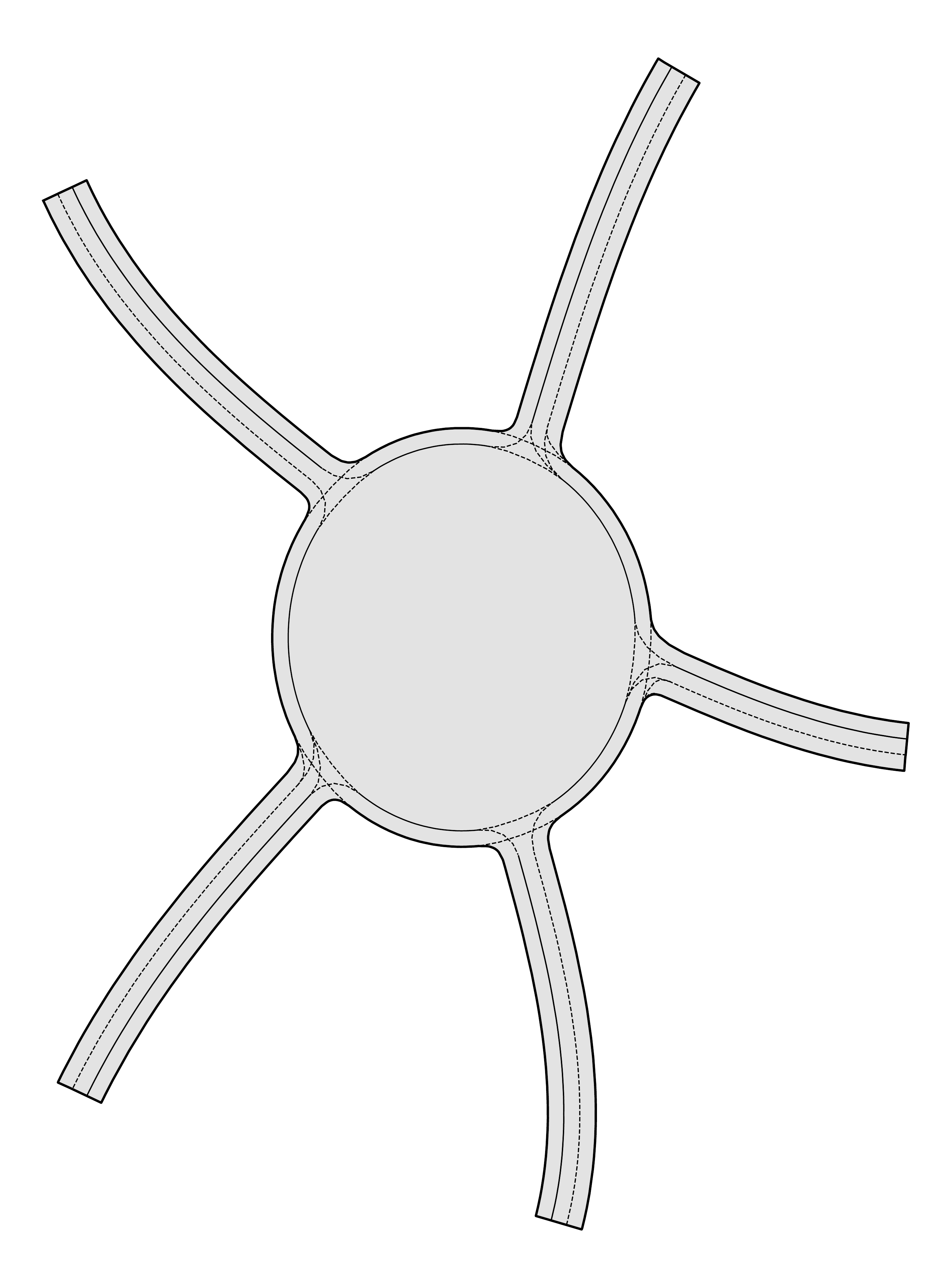}
        \includegraphics[width=0.25\textwidth]{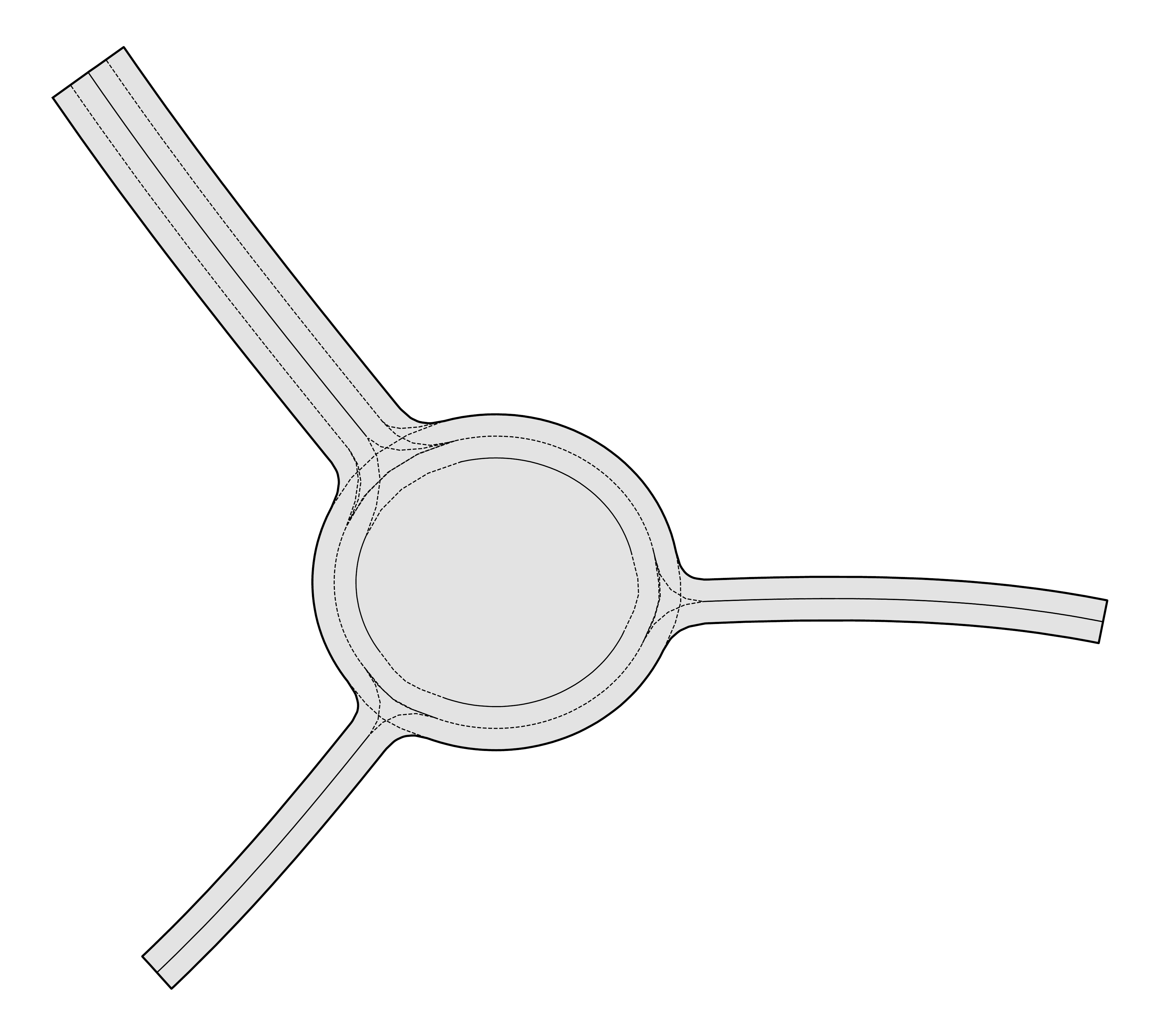}
        \includegraphics[width=0.35\textwidth]{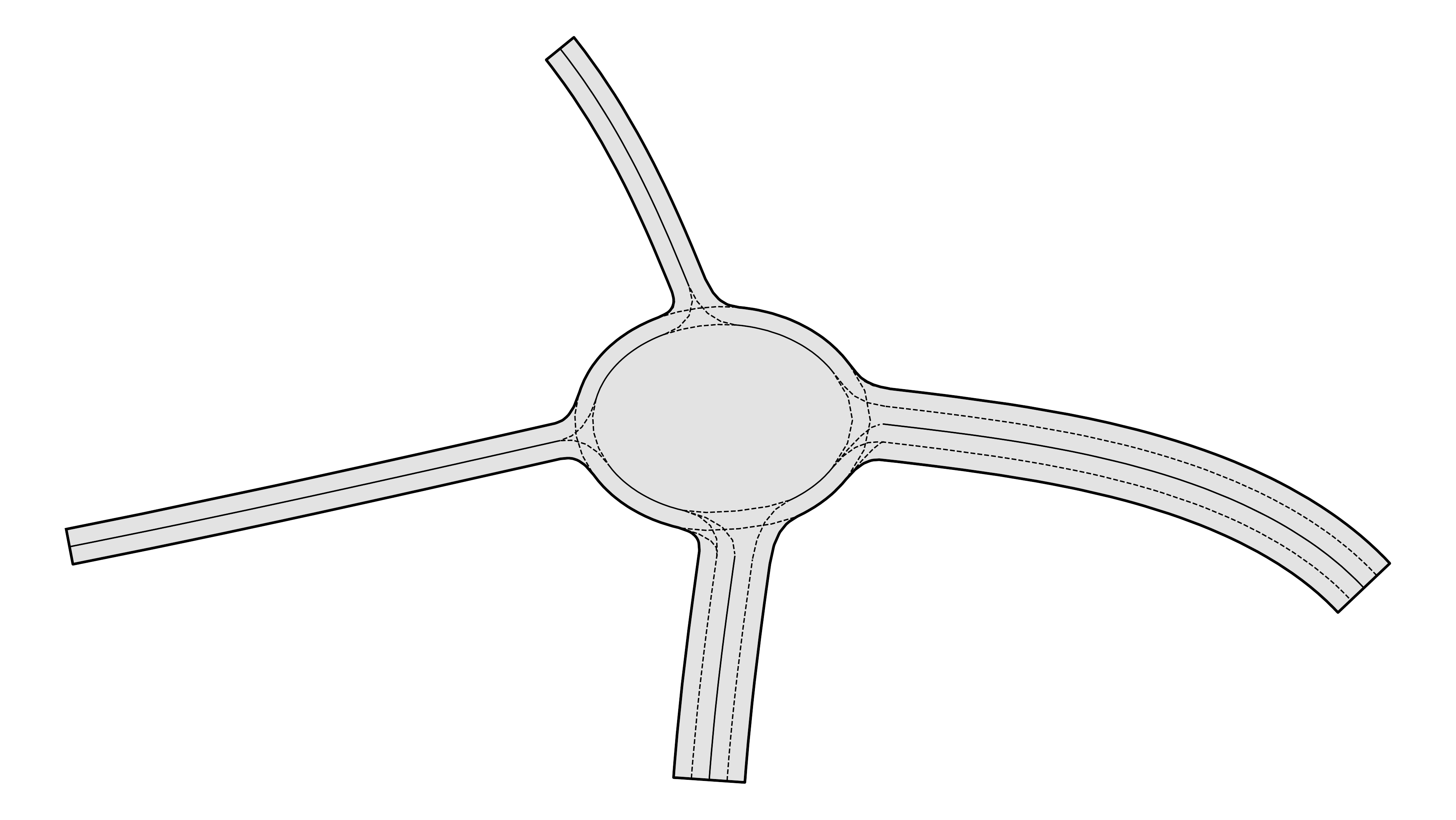}
    \label{subfig:map-vis-roundabout}
    }
    
    \subfigure[Highway-Entry]{
        \includegraphics[width=0.2\textwidth]{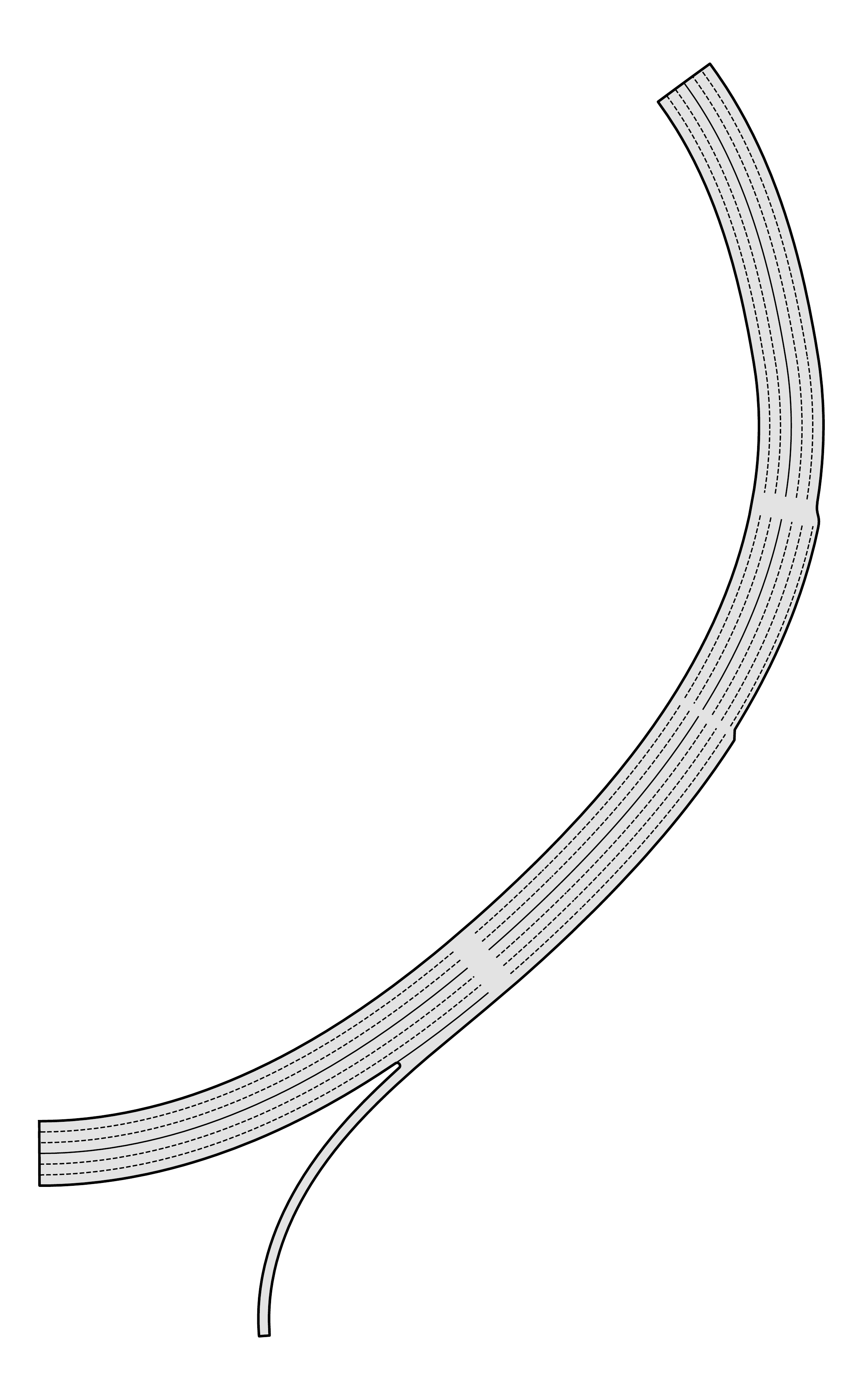}
        \includegraphics[width=0.5\textwidth]{imgs/generated_maps/highway_entry/12.pdf}
        \includegraphics[width=0.3\textwidth]{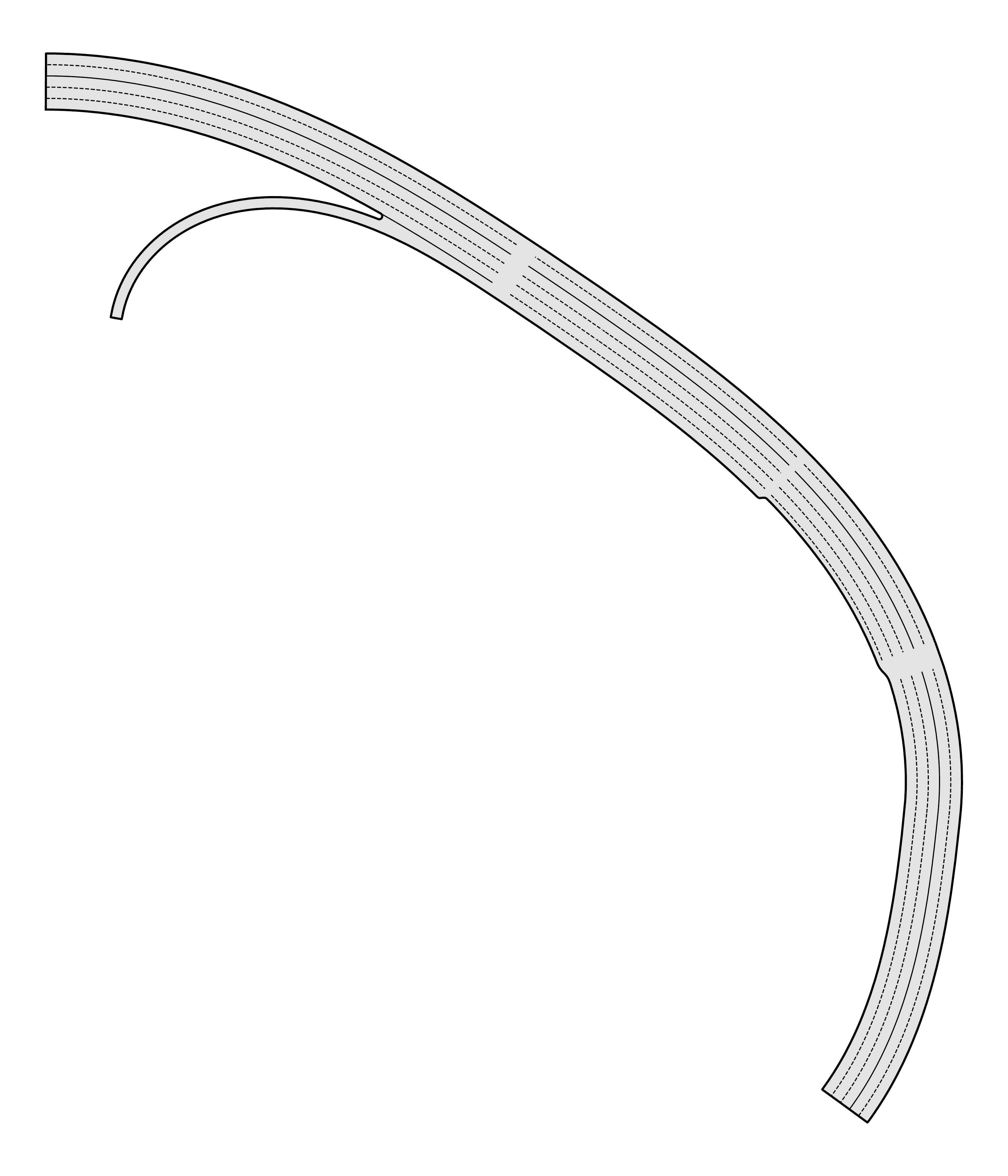}
    \label{subfig:map-vis-entry}
    }
    
    \subfigure[Highway-Drive]{            
        \includegraphics[width=0.4\textwidth]{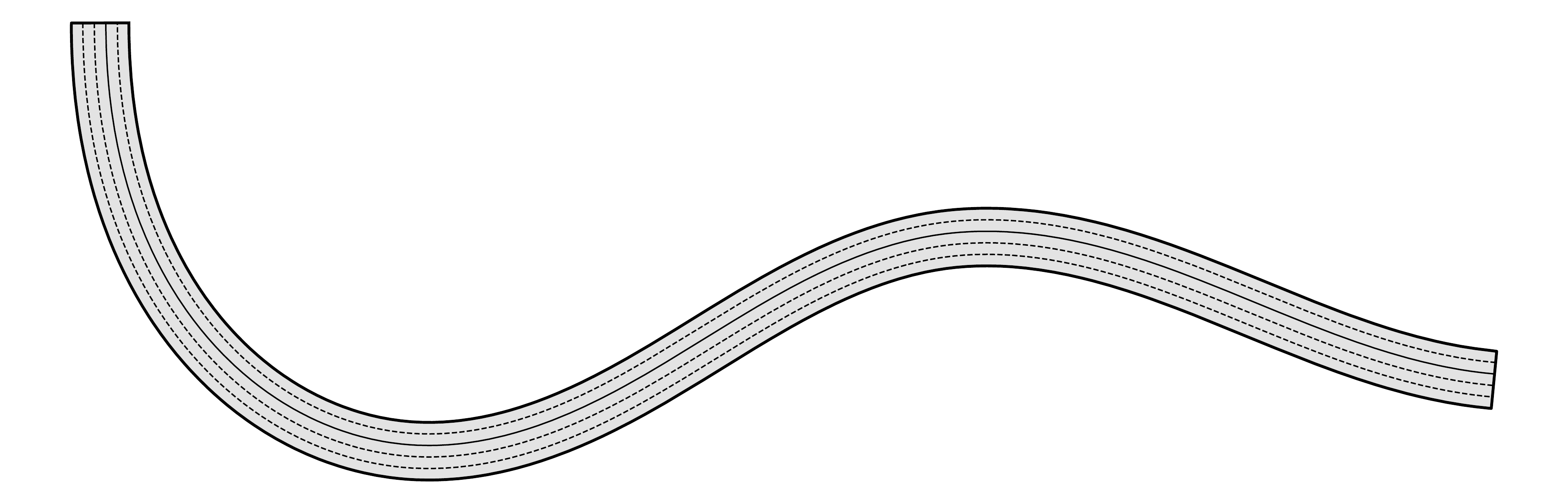}
        \includegraphics[width=0.4\textwidth]{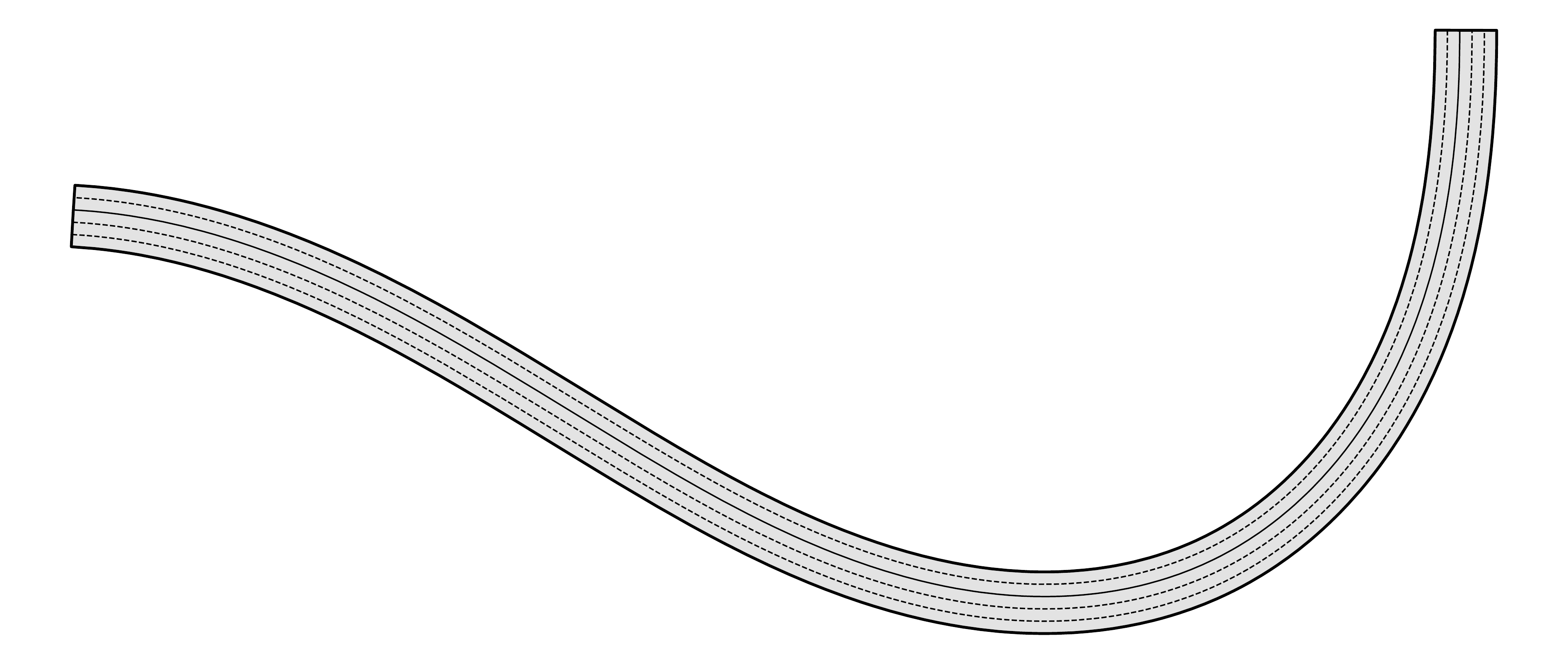}
    \label{subfig:map-vis-drive}
    }
    
    \subfigure[Highway-Exit]{            
        \includegraphics[width=0.3\textwidth]{imgs/generated_maps/highway_exit/10.pdf}
        \includegraphics[width=0.3\textwidth]{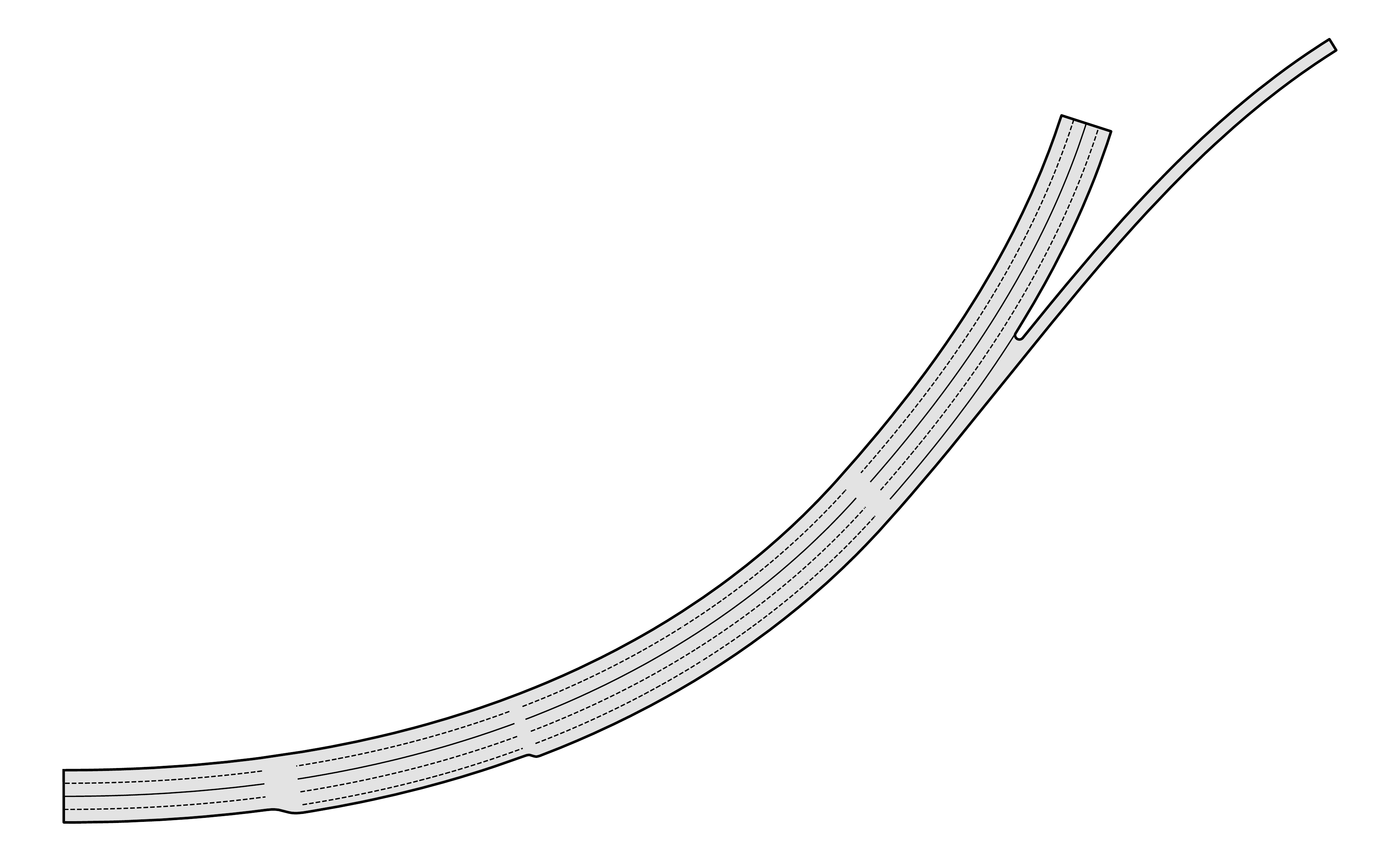}
        \includegraphics[width=0.4\textwidth]{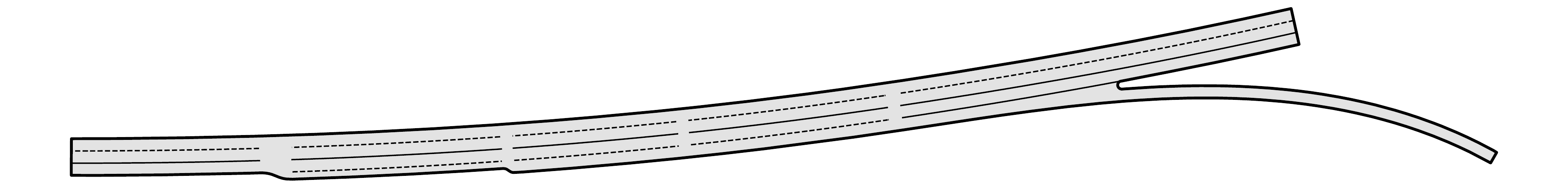}
    \label{subfig:map-vis-exit}
    }
    \caption{Selection of generated maps of the Driver Dojo core scenarios.}
    \label{fig:map-vis}
\end{figure}

\begin{figure}
    \centering
    \subfigure[Intersection]{
    \includegraphics[width=0.9\textwidth]{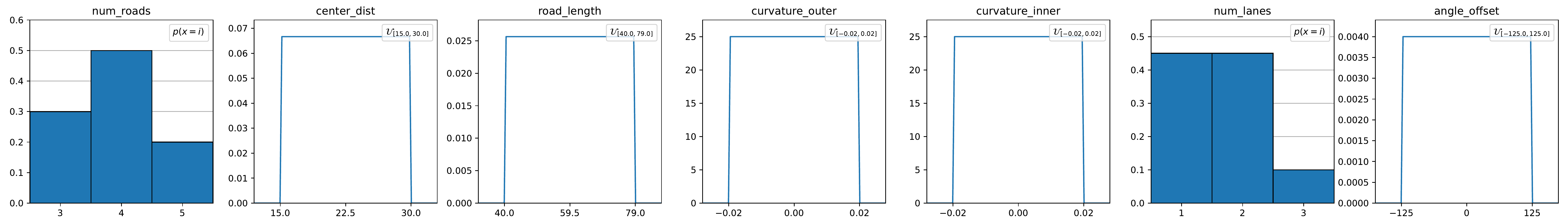}
    \label{subfig:map-intersection-dist}
    }
    
    \subfigure[Roundabout]{
    \includegraphics[width=0.9\textwidth]{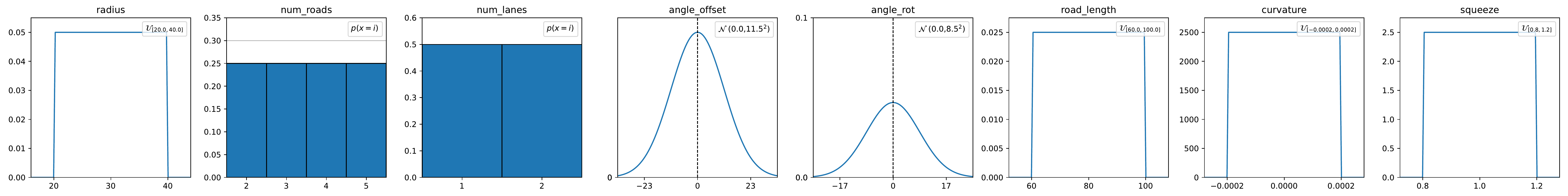}
    \label{subfig:map-roundabout-dist}
    }    
    
    \subfigure[Highway-Drive]{
    \includegraphics[width=0.7\textwidth]{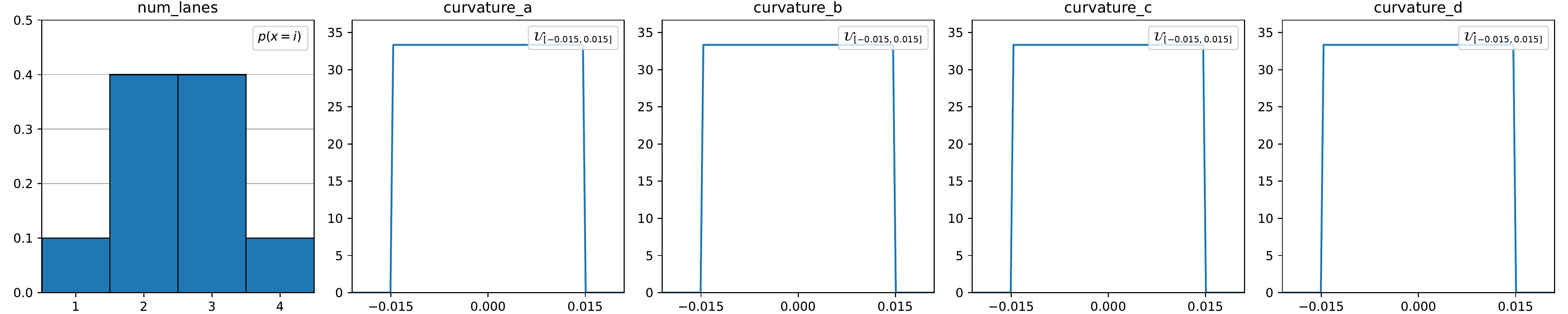}
    \label{subfig:map-drive-dist}
    }
    
    \subfigure[Highway-Entry]{
    \includegraphics[width=0.7\textwidth]{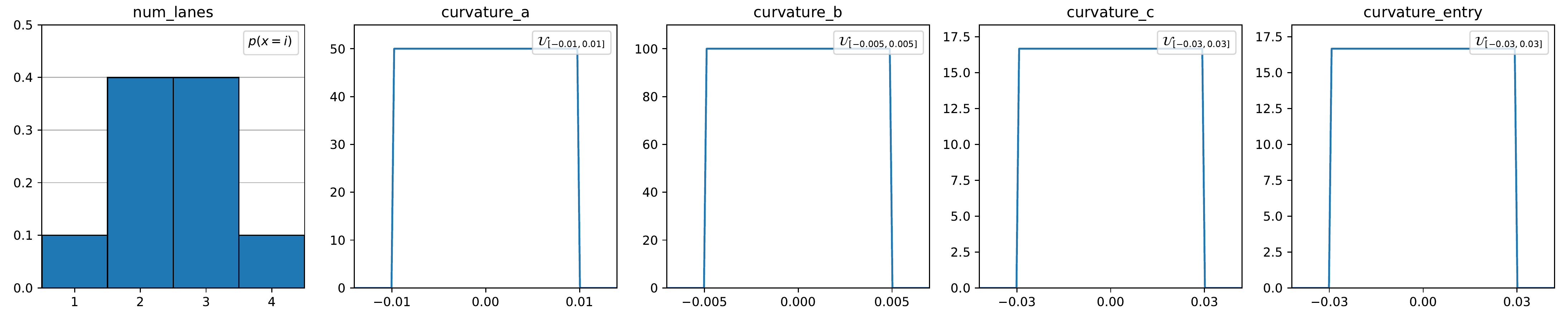}
    \label{subfig:map-entry-dist}
    }
    
    \subfigure[Highway-Exit]{
    \includegraphics[width=0.7\textwidth]{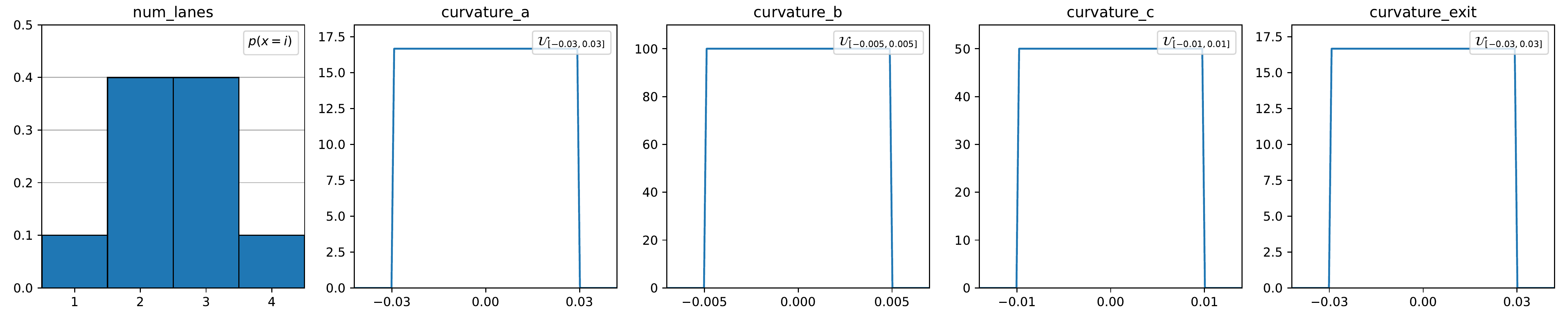}
    \label{subfig:map-exit-dist}
    }
    \caption{Sampling distributions for free parameters of the core scenarios provided by Driver Dojo.}
    \label{fig:map-dist}
\end{figure}

\begin{figure}
    \centering
    \subfigure[Intersection]{
    \includegraphics[width=0.6\textwidth]{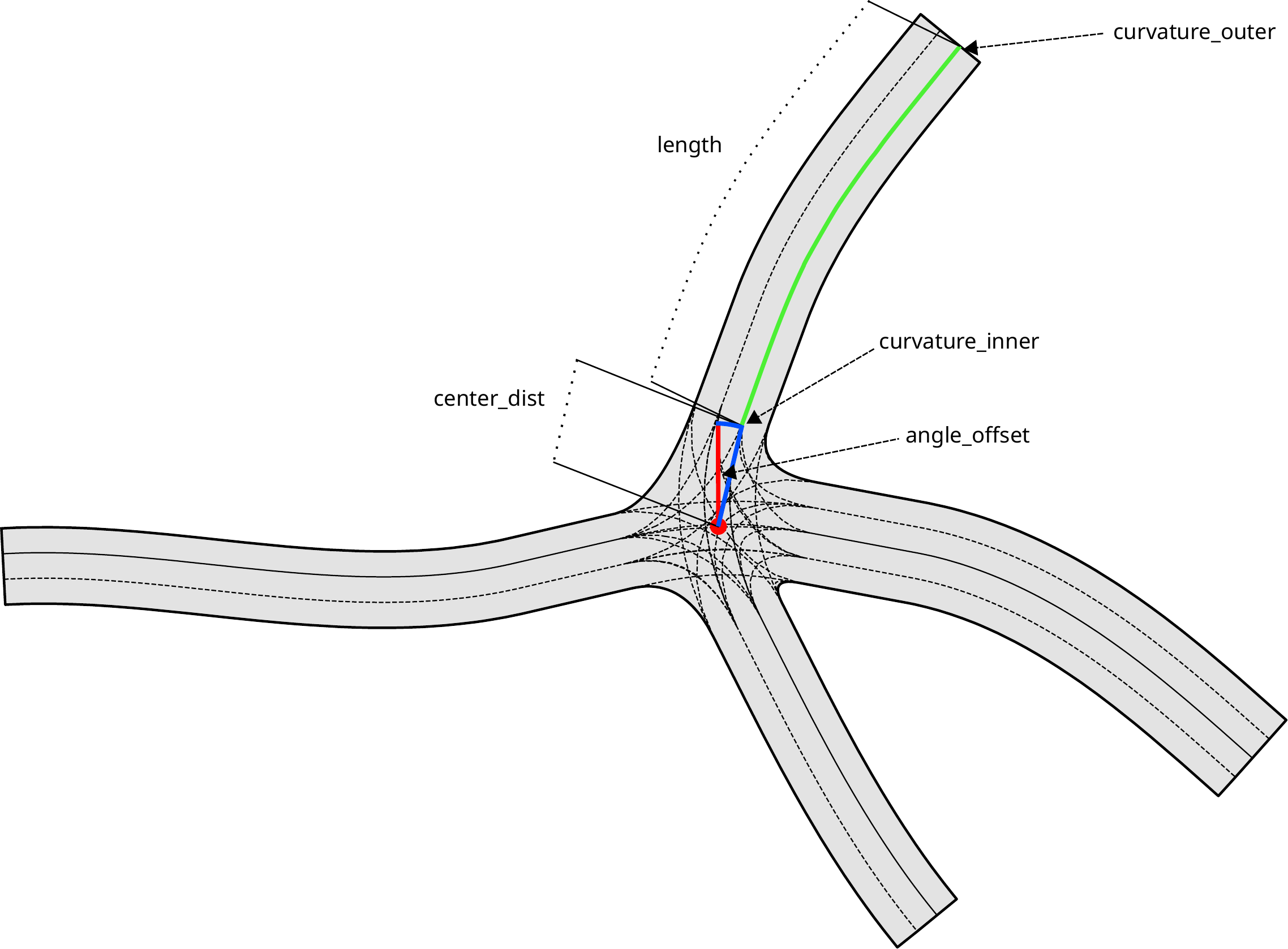}
    \label{subfig:map-intersection-method}
    }
         
    \subfigure[Roundabout]{
    \includegraphics[width=0.7\textwidth]{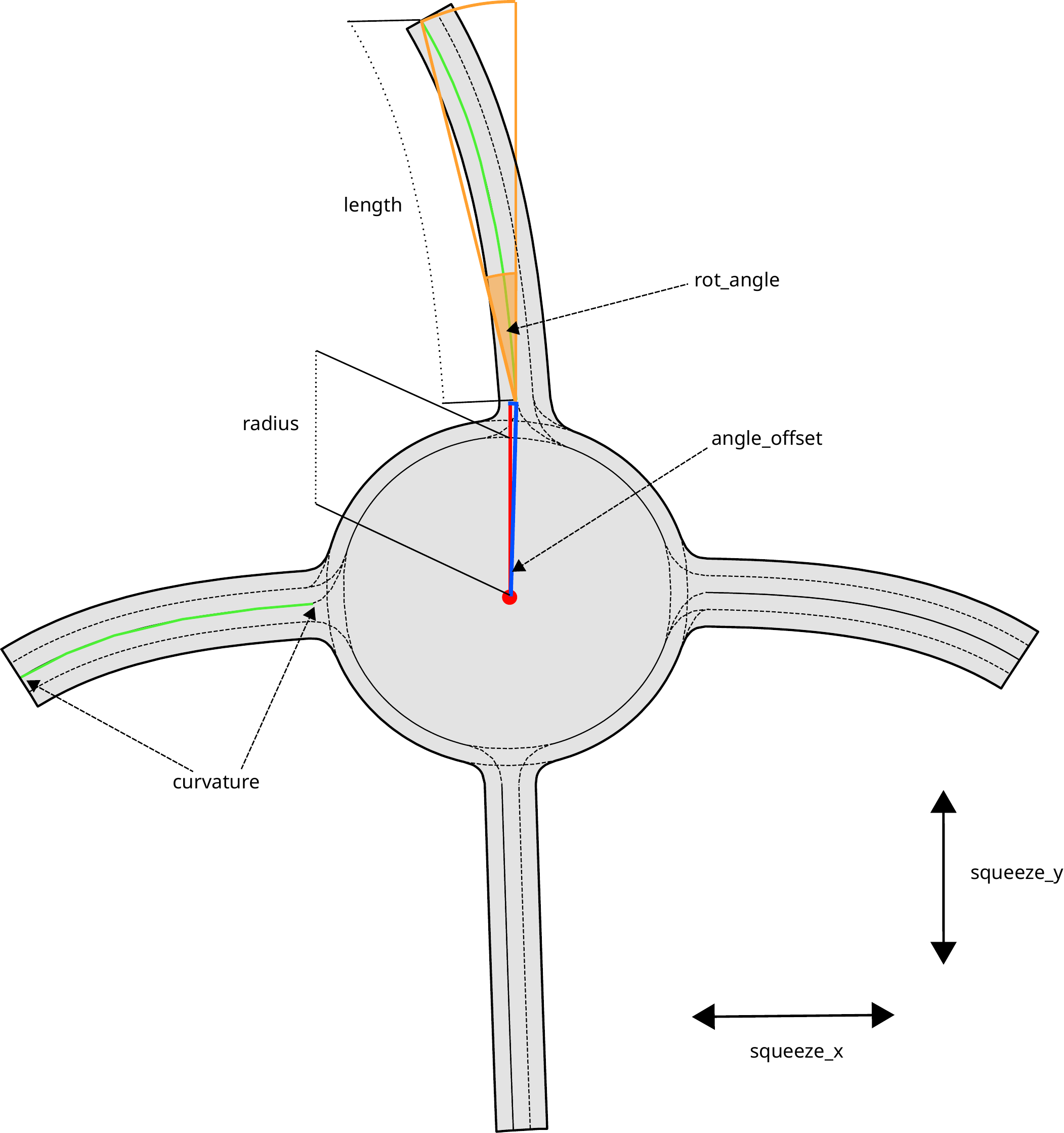}
    \label{subfig:map-roundabout-method}
    }
    \caption{Visualization of free parameters for the Intersection and Roundabout scenario.}
    \label{fig:map-method}
\end{figure}

Similar as in the previous section, we show distributions for free street network parameters for every scenario in Figure~\ref{fig:map-dist} and give visualizations aiding our explanation in Figure~\ref{fig:map-method}. Figure~\ref{fig:map-vis} illustrates a selection of generated maps for every scenario.

For the Intersection scenario (Figure~\ref{subfig:map-intersection-dist} and Figure~\ref{subfig:map-intersection-method}) we limited ourselves to intersections with three, four or five incoming lanes, as these are the most common ones, and weigh their occurrence with slightly different probabilities. For every incoming lane, we independently sample the number of right and left lanes (one, two or three lanes), as well as the distance of the respective road endpoint to the junction center and its length. We place every lane at an angle $\frac{a * 2\pi }{\text{\textit{num\_lanes}}} + \text{\textit{lane\_offset}}_a$, where $a$ is the index of the lane and $\text{\textit{lane\_offset}}_a$ is sampled from a Gaussian distribution with zero mean. As for all road shapes in our scenarios, we use clothoids parameterized through the road length, a start curvature and an end curvature. For the Intersection scenario, we draw independent start and end curvature values.

Moving on the Roundabout scenario (Figure~\ref{subfig:map-roundabout-dist} and Figure~\ref{subfig:map-roundabout-method}), we allow for five incoming lanes and give every possibility equal probabilities instead. Roads are assigned at an equal distance from the roundabout center which we compensate for squeezing the whole network structure with two independent values drawn from a uniform distribution. The number of lanes inside the roundabout is constrained to one and two lanes. In contrast to the Intersection scenario, we use the same curvature values for the start and endpoint of incoming roads. However, in addition to what has been done in the Intersection scenario, we displace the endpoint of every incoming lane based on a second angular value.

For the Highway scenarios (Figure ~\ref{subfig:map-drive-dist}, Figure ~\ref{subfig:map-entry-dist} and Figure ~\ref{subfig:map-exit-dist}), we divide the whole highway section into parts of roughly equal size ($100$\si{m}-$200$\si{m} in general), which we hold fixed and do not randomize in this case, and use successive clothoids with varying curvatures to inject variation into the scenarios. For the Highway-Entry scenario, we constrain the curvature of the entry road clothoid to $c_0 - 0.01$, where $c_0$ is the curvature at the beginning of the highway section, in order to avoid overlapping. The same holds true for the Highway-Exit scenario, where we constrain the exit lane with respect to the end curvature of the section.

\subsection{Experimental Setup}
\label{sec:setup}
In this section we present our experimental setup, including RL and software frameworks used for training and evaluation, algorithm hyperparameters and our compute resources. We give a full documentation of how to reproduce our results inside the \texttt{README.md} file of our GitHub repository.
\begin{table}
\centering
\caption{Hyperparameters of RL methods (DQN, FQF and PPO) used to produce the results of our experimental section.}
\label{table:hyperparams}
\tiny{
\subfigure[DQN]{
    \begin{tabular}{ll}
    \toprule
    lr & $0.0003$ \\
    lr\_decay & \texttt{False} \\
    gamma & $0.99$ \\
    n\_step & $3$ \\
    buffer\_size & $100000$ \\
    target\_update\_freq & $1000$ \\
    batch\_size & $64$ \\
    eps\_train & $1.0$ \\
    eps\_test & $0.001$ \\
    eps\_fraction & $0.2$ \\
    hidden\_sizes & $[256, 256]$ \\
    duelling\_q\_sizes & $[256, 256]$ \\
    duelling\_v\_sizes & $[256, 256]$ \\
    reward normalization & \texttt{False} \\
    max\_grad\_norm & \texttt{False} \\
    prioritized\_replay & \texttt{False} \\
    is\_double & \texttt{True} \\
    alpha & 0.6 \\
    beta & 0.4 \\
    steps\_per\_collect & $8$ \\
    parallel environments & $8$ \\
    timesteps & $1000000$ \\
    \bottomrule
    \end{tabular}
    \label{table:hyperparams-dqn}
}
\subfigure[FQF]{
    \begin{tabular}{ll}
    \toprule
    lr & $0.0003$ \\
    lr\_decay & \texttt{False} \\
    fraction\_lr & 2.5e-9 \\
    gamma & $0.99$ \\
    num\_fractions & 32 \\
    num\_cosines & 64 \\
    ent\_coef & 10.0 \\
    n\_step & 3 \\
    buffer\_size & $100000$ \\
    target\_update\_freq & $1000$ \\
    batch\_size & $64$ \\
    eps\_train & $1.0$ \\
    eps\_test & $0.001$ \\
    eps\_fraction & $0.2$ \\
    hidden\_sizes & $[256, 256]$ \\
    reward normalization & \texttt{False} \\
    max\_grad\_norm & \texttt{False} \\
    prioritized\_replay & \texttt{False} \\
    alpha & 0.6 \\
    beta & 0.4 \\
    steps\_per\_collect & $8$ \\
    parallel environments & $8$ \\
    timesteps & $1000000$ \\
    \bottomrule
    \end{tabular}
    \label{table:hyperparams-fqf}
}
\subfigure[PPO]{
    \begin{tabular}{ll}
    \toprule
    lr & $0.0003$ \\
    lr\_decay & \texttt{False} \\
    gamma & $0.99$ \\
    gae\_lambda & $0.95$ \\
    batch\_size & $256$ \\
    hidden\_sizes & $[256, 256]$ \\
    epsilon\_clip & $0.2$ \\
    value\_clip & \texttt{False} \\
    advantage\_norm & \texttt{True} \\
    reward\_norm & \texttt{False} \\
    vf\_coef & $0.5$ \\
    max\_grad\_norm & \texttt{False} \\
    update\_steps & $5$ \\
    steps\_per\_collect & $2048$ \\
    parallel environments & $8$ \\
    timesteps & $1000000$ \\
    \bottomrule
    \end{tabular}
    \label{table:hyperparams-ppo}
}}
\end{table}

\textbf{Frameworks and Hyperparameters.} We used RL implementations provided by the Tianshou~\cite{weng2021tianshou} framework and further made use of Hydra~\cite{Yadan2019Hydra}, a framework that allows for composable experiment configurations and easy reproducability of our experiments.

The hyperparameters we used are shown in Table~\ref{table:hyperparams}. We aimed to follow common hyperparameter settings from the literature, while also allowing for fair comparisson between the algorithms, for example, regarding number of training steps, neutral network sizes and learning rate, as well as other parameters. For models trained on visual observations, we used the architecture proposed by Mnih et al.~\cite{mnih2015human} as feature extractor. In detail, we used $32$ $8 \times 8 \times 3$, 64 $4 \times 4 \times 32$, and and 64 $3 \times 3 \times 64$ convolutional filter layers with strides $4$, $3$ and $1$, respectively, and ReLU activations after every layer. The flattened outputs are then fed into a feed-forward layer network with the same architecture as described in Table~\ref{table:hyperparams}.

\textbf{Compute Resources.} We used three different machine types to run our experiments. First, we used a personal workstation with an AMD Ryzen 9 5900X 12-Core processor and an Nvidia RTX 3090 to run exploratory experiments and train a handful of models appearing in the experimental results. Second, we used a HPC cluster, where some models were trained on CPU-exclusive nodes with Intel Xeon Gold 5120 CPU @ 2.20GHz and, especially vision-based models, on GPU nodes with equivalent CPU resources but four additional Nvidia Tesla V100 GPUs each. The desktop workstation has 64 GB, the CPU nodes 502GB and the GPU nodes 187 GB of main memory. In total, we roughly occupied three HPC nodes and ran the desktop workstation at full capacity for 7 days to train all of the models appearing in our experimental section.

\subsection{Supplementary Results}
\label{sec:results}
\begin{table}[b!]
\caption{Crash rate (CrR) and completion rate (CoR) for agents trained on the Intersection and Highway-Entry scenario using the TPS action space.}
\label{table:experiment2_rates-sup}
\centering
\subfigure[Intersection]{
\begin{tabular}{lll}
\toprule
Agent & CrR & CoR \\
\midrule
DQN-100 & 21.25 & 59.06 \\
DQN-10000 & 53.88 & 34.87 \\
PPO-100 & 23.87 & 56.06 \\
PPO-10000 & 34.71 & 43.63 \\
FQF-100 & 16.25 & 55.91\\
FQF-10000 & 18.48 & 51.20 \\
\bottomrule
\end{tabular}
}
\subfigure[Highway-Entry]{
\begin{tabular}{lll}
\toprule
Agent & CrR & CoR \\
\midrule
DQN-100 & 1.9 & 83.14 \\
DQN-10000 & 6.41 & 46.45  \\
PPO-100 & 14.56 & 75.14 \\
PPO-10000 & 18.06 & 70.68 \\
FQF-100 & 10.49 & 29.92 \\
FQF-10000 & 11.44 & 31.81 \\
\bottomrule
\end{tabular}
}
\end{table}
In this section, we give crash rates and completion rates we skipped in the main text. Results are listed in Table~\ref{table:experiment2_rates-sup}.

\clearpage
\bibliographystyle{IEEEtran}
\bibliography{references}

\end{document}